# Value-Function Approximations for Partially Observable Markov Decision Processes

**Milos Hauskrecht**    MILOS@CS.BROWN.EDU
*Computer Science Department, Brown University*
*Box 1910, Brown University, Providence, RI 02912, USA*

## Abstract

Partially observable Markov decision processes (POMDPs) provide an elegant mathematical framework for modeling complex decision and planning problems in stochastic domains in which states of the system are observable only indirectly, via a set of imperfect or noisy observations. The modeling advantage of POMDPs, however, comes at a price — exact methods for solving them are computationally very expensive and thus applicable in practice only to very simple problems. We focus on efficient approximation (heuristic) methods that attempt to alleviate the computational problem and trade off accuracy for speed. We have two objectives here. First, we survey various approximation methods, analyze their properties and relations and provide some new insights into their differences. Second, we present a number of new approximation methods and novel refinements of existing techniques. The theoretical results are supported by experiments on a problem from the agent navigation domain.

## 1. Introduction

Making decisions in dynamic environments requires careful evaluation of the cost and benefits not only of the immediate action but also of choices we may have in the future. This evaluation becomes harder when the effects of actions are stochastic, so that we must pursue and evaluate many possible outcomes in parallel. Typically, the problem becomes more complex the further we look into the future. The situation becomes even worse when the outcomes we can observe are imperfect or unreliable indicators of the underlying process and special actions are needed to obtain more reliable information. Unfortunately, many real-world decision problems fall into this category.

Consider, for example, a problem of patient management. The patient comes to the hospital with an initial set of complaints. Only rarely do these allow the physician (decision-maker) to diagnose the underlying disease with certainty, so that a number of disease options generally remain open after the initial evaluation. The physician has multiple choices in managing the patient. He/she can choose to do nothing (wait and see), order additional tests and learn more about the patient state and disease, or proceed to a more radical treatment (e.g. surgery). Making the right decision is not an easy task. The disease the patient suffers can progress over time and may become worse if the window of opportunity for a particular effective treatment is missed. On the other hand, selection of the wrong treatment may make the patient's condition worse, or may prevent applying the correct treatment later. The result of the treatment is typically non-deterministic and more outcomes are possible. In addition, both treatment and investigative choices come with different costs. Thus, in a course of patient management, the decision-maker must carefully evaluate the costs and benefits of both current and future choices, as well as their interaction and ordering. Other decision problems with similar characteristics — complex temporal cost-benefit tradeoffs, stochasticity, and partial observability of the underlying controlled process — include robot navigation, target tracking, machine mantainance and replacement, and the like.

Sequential decision problems can be modeled as *Markov decision processes (MDPs)* (Bellman, 1957; Howard, 1960; Puterman, 1994; Boutilier, Dean, & Hanks, 1999) and their extensions. The model of choice for problems similar to patient management is the *partially observable Markov decision process (POMDP)* (Drake, 1962; Astrom, 1965; Sondik, 1971; Lovejoy, 1991b). The POMDP represents two sources of uncertainty: stochasticity of the underlying controlled process (e.g. disease dynamics in the patient management problem), and imperfect observability of its states via a set of noisy observations (e.g. symptoms, findings, results of tests). In addition, it lets us model in a uniform way both control and information-gathering (investigative) actions, as well as their effects and cost-benefit trade-offs. Partial observability and the ability to model and reason with information-gathering actions are the main features that distinguish the POMDP from the widely known *fully observable Markov decision process* (Bellman, 1957; Howard, 1960).

Although useful from the modeling perspective, POMDPs have the disadvantage of being hard to solve (Papadimitriou & Tsitsiklis, 1987; Littman, 1996; Mundhenk, Goldsmith, Lusena, & Allender, 1997; Madani, Hanks, & Condon, 1999), and optimal or $\epsilon$-optimal solutions can be obtained in practice only for problems of low complexity. A challenging goal in this research area is to exploit additional structural properties of the domain and/or suitable approximations (heuristics) that can be used to obtain good solutions more efficiently.

We focus here on heuristic approximation methods, in particular approximations based on value functions. Important research issues in this area are the design of new and efficient algorithms, as well as a better understanding of the existing techniques and their relations, advantages and disadvantages. In this paper we address both of these issues. First, we survey various value-function approximations, analyze their properties and relations and provide some insights into their differences. Second, we present a number of new methods and novel refinements of existing techniques. The theoretical results and findings are also supported empirically on a problem from the agent navigation domain.

## 2. Partially Observable Markov Decision Processes

A *partially observable Markov decision process (POMDP)* describes a stochastic control process with partially observable (hidden) states. Formally, it corresponds to a tuple $(S, A, \Theta, T, O, R)$ where $S$ is a set of states, $A$ is a set of actions, $\Theta$ is a set of observations, $T : S \times A \times S \to [0, 1]$ is a set of transition probabilities that describe the dynamic behavior of the modeled environment, $O : S \times A \times \Theta \to [0, 1]$ is a set of observation probabilities that describe the relationships among observations, states and actions, and $R : S \times A \times S \to \mathbb{R}$ denotes a reward model that assigns rewards to state transitions and models payoffs associated with such transitions. In some instances the definition of a POMDP also includes an *a priori* probability distribution over the set of initial states $S$.

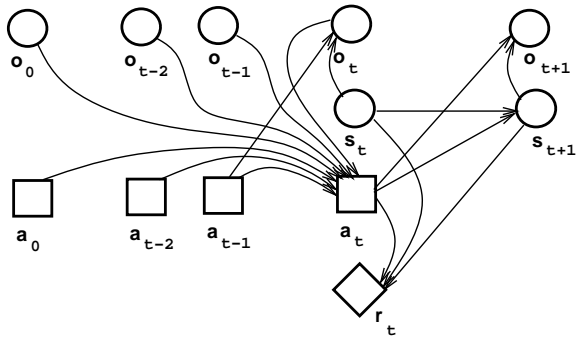

Figure 1: Part of the influence diagram describing a POMDP model. Rectangles correspond to decision nodes (actions), circles to random variables (states) and diamonds to reward nodes. Links represent the dependencies among the components. $s_t, a_t, o_t$ and $r_t$ denote state, action, observation and reward at time $t$. Note that an action at time $t$ depends only on past observations and actions, not on states.

## 2.1 Objective Function

Given a POMDP, the goal is to construct a *control policy* that maximizes an *objective (value) function*. The objective function combines partial (stepwise) rewards over multiple steps using various kinds of decision nodes. Typically, the models are cumulative and based on expectations. Two models are frequently used in practice:

- a *finite-horizon* model in which we maximize $E(\sum_{t=0}^{T} r_t)$, where $r_t$ is a reward obtained at time $t$.

- an *infinite-horizon discounted* model in which we maximize $E(\sum_{t=0}^{\infty} \gamma^t r_t)$, where $0 < \gamma < 1$ is a discount factor.

Note that POMDPs and cumulative decision models provide a rich language for modeling various control objectives. For example, one can easily model goal-achievement tasks (a specific goal must be reached) by giving a large reward for a transition to that state and zero or smaller rewards for other transitions.

In this paper we focus primarily on discounted infinite-horizon model. However, the results can be easily applied also to the finite-horizon case.

## 2.2 Information State

In a POMDP the process states are hidden and we cannot observe them while making a decision about the next action. Thus, our action choices are based only on the information available to us or on quantities derived from that information. This is illustrated in the influence diagram in Figure 1, where the action at time $t$ depends only on previous observations and actions, not on states. Quantities summarizing all information are called *information states*. *Complete information states* represent a trivial case.

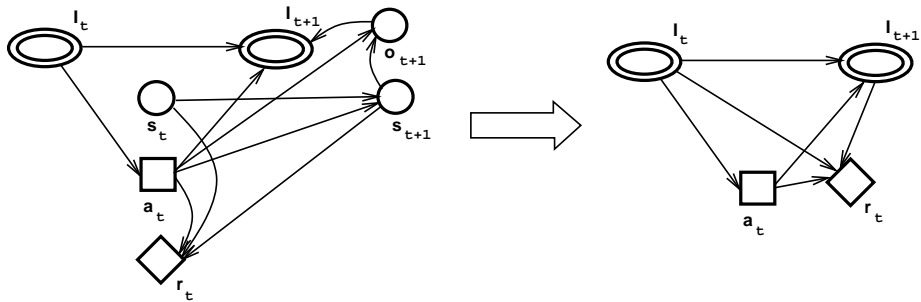

Figure 2: Influence diagram for a POMDP with information states and corresponding information-state MDP. Information states ($I_t$ and $I_{t+1}$) are represented by double-circled nodes. An action choice (rectangle) depends only on the current information state.

**Definition 1** *(Complete information state). The complete information state at time t (denoted $I_t^C$) consists of:*

- *a prior belief $b_0$ on states in $S$ at time 0;*

- *a complete history of actions and observations $\{o_0, a_0, o_1, a_1, \cdots, o_{t-1}, a_{t-1}, o_t\}$ starting from time $t = 0$.*

A sequence of information states defines a controlled Markov process that we call an *information-state Markov decision process* or *information-state MDP*. The policy for the information-state MDP is defined in terms of a control function $\mu : \mathcal{I} \to A$ mapping information state space to actions. The new information state ($I_t$) is a deterministic function of the previous state ($I_{t-1}$), the last action ($a_{t-1}$) and the new observation ($o_t$):

$$I_t = \tau(I_{t-1}, o_t, a_{t-1}).$$

$\tau : \mathcal{I} \times \Theta \times A \to \mathcal{I}$ is the *update function* mapping the information state space, observations and actions back to the information space.[1] It is easy to see that one can always convert the original POMDP into the information-state MDP by using complete information states. The relation between the components of the two models and a sketch of a reduction of a POMDP to an information-state MDP, are shown in Figure 2.

### 2.3 Bellman Equations for POMDPs

An information-state MDP for the infinite-horizon discounted case is like a fully-observable MDP and satisfies the standard fixed-point (Bellman) equation:

$$V^*(I) = \max_{a \in A} \left\{ \rho(I, a) + \gamma \sum_{I'} P(I'|I, a) V^*(I') \right\}. \tag{1}$$

---

1. In this paper, $\tau$ denotes the generic update function. Thus we use the same symbol even if the information state space is different.

Here, $V^*(I)$ denotes the optimal value function maximizing $E(\sum_{t=0}^{\infty} \gamma^t r_t)$ for state $I$. $\rho(I, a)$ is the expected one-step reward and equals

$$\rho(I, a) = \sum_{s \in S} \rho(s, a) P(s|I) = \sum_{s \in S} \sum_{s' \in S} R(s, a, s') P(s'|s, a) P(s|I).$$

$\rho(s, a)$ denotes an expected one-step reward for state $s$ and action $a$.

Since the next information state $I' = \tau(I, o, a)$ is a deterministic function of the previous information state $I$, action $a$, and the observation $o$, the Equation 1 can be rewritten more compactly by summing over all possible observations $\Theta$:

$$V^*(I) = \max_{a \in A} \left\{ \sum_{s \in S} \rho(s, a) P(s|I) + \gamma \sum_{o \in \Theta} P(o|I, a) V^*(\tau(I, o, a)) \right\}. \tag{2}$$

The optimal policy (control function) $\mu^* : \mathcal{I} \to A$ selects the value-maximizing action

$$\mu^*(I) = \arg\max_{a \in A} \left\{ \sum_{s \in S} \rho(s, a) P(s|I) + \gamma \sum_{o \in \Theta} P(o|I, a) V^*(\tau(I, o, a)) \right\}. \tag{3}$$

The value and control functions can be also expressed in terms of *action-value functions (Q-functions)*

$$V^*(I) = \max_{a \in A} Q^*(I, a) \qquad \mu^*(I) = \arg\max_{a \in A} Q^*(I, a),$$
$$Q^*(I, a) = \sum_{s \in S} \rho(s, a) P(s|I) + \gamma \sum_{o \in \Theta} P(o|I, a) V^*(\tau(I, o, a)). \tag{4}$$

A *Q*-function corresponds to the expected reward for chosing a fixed action ($a$) in the first step and acting optimally afterwards.

### 2.3.1 SUFFICIENT STATISTICS

To derive Equations 1—3 we implicitly used complete information states. However, as remarked earlier, the information available to the decision-maker can be also summarized by other quantities. We call them *sufficient information states*. Such states must preserve the necessary information content and also the Markov property of the information-state decision process.

**Definition 2** (*Sufficient information state process*). *Let $\mathcal{I}$ be an information state space and $\tau : \mathcal{I} \times A \times \Theta \to \mathcal{I}$ be an update function defining an information process $I_t = \tau(I_{t-1}, a_{t-1}, o_t)$. The process is sufficient with regard to the optimal control when, for any time step $t$, it satisfies*

$$P(s_t|I_t) = P(s_t|I_t^C)$$
$$P(o_t|I_{t-1}, a_{t-1}) = P(o_t|I_{t-1}^C, a_{t-1}),$$

*where $I_t^C$ and $I_{t-1}^C$ are complete information states.*

It is easy to see that Equations 1 — 3 for complete information states must hold also for sufficient information states. The key benefit of sufficient statistics is that they are often easier to manipulate and store, since unlike complete histories, they may not expand with time. For example, in the standard POMDP model it is sufficient to work with belief states that assign probabilities to every possible process state (Astrom, 1965).[2] In this case the Bellman equation reduces to:

$$V(b) = \max_{a \in A} \left\{ \sum_{s \in S} \rho(s, a)b(s) + \gamma \sum_{o \in \Theta} \sum_{s \in S} P(o|s, a)b(s)V(\tau(b, o, a)) \right\}, \quad (5)$$

where the next-step belief state $b'$ is

$$b'(s) = \tau(b, o, a)(s) = \beta P(o|s, a) \sum_{s' \in S} P(s|a, s')b(s').$$

$\beta = 1/P(o|b, a)$ is a normalizing constant. This defines a *belief-state MDP* which is a special case of a continuous-state MDP. Belief-state MDPs are also the primary focus of our investigation in this paper.

### 2.3.2 VALUE-FUNCTION MAPPINGS AND THEIR PROPERTIES

The Bellman equation 2 for the belief-state MDP can be also rewritten in the value-function mapping form. Let $\mathcal{V}$ be a space of real-valued bounded functions $V : \mathcal{I} \to \mathbb{R}$ defined on the belief information space $\mathcal{I}$, and let $h : \mathcal{I} \times A \times B \to \mathbb{R}$ be defined as

$$h(b, a, V) = \sum_{s \in S} \rho(s, a)b(s) + \gamma \sum_{o \in \Theta} \sum_{s \in S} P(o|s, a)b(s)V(\tau(b, o, a)).$$

Now by defining the value function mapping $H : \mathcal{V} \to \mathcal{V}$ as $(HV)(b) = \max_{a \in A} h(b, a, V)$, the Bellman equation 2 for all information states can be written as $V^* = HV^*$. It is well known that $H$ (for MDPs) is an isotone mapping and that it is a contraction under the supremum norm (see (Heyman & Sobel, 1984; Puterman, 1994)).

**Definition 3** *The mapping $H$ is isotone, if $V, U \in \mathcal{V}$ and $V \leq U$ implies $HV \leq HU$.*

**Definition 4** *Let $\|.\|$ be a supremum norm. The mapping $H$ is a contraction under the supremum norm, if for all $V, U \in \mathcal{V}$, $\|HV - HU\| \leq \beta \|V - U\|$ holds for some $0 \leq \beta < 1$.*

## 2.4 Value Iteration

The optimal value function (Equation 2) or its approximation can be computed using *dynamic programming* techniques. The simplest approach is the *value iteration* (Bellman, 1957) shown in Figure 3. In this case, the optimal value function $V^*$ can be determined in the limit by performing a sequence of value-iteration steps $V_i = HV_{i-1}$, where $V_i$ is the $i$th approximation of the value function ($i$th value function).[3] The sequence of estimates

---

2. Models in which belief states are not sufficient include POMDPs with observation and action channel lags (see Hauskrecht (1997)).

3. We note that the same update $V_i = HV_{i-1}$ can be applied to solve the finite-horizon problem in a standard way. The difference is that $V_i$ now stands for the $i$-steps-to-go value function and $V_0$ represents the value function (rewards) for end states.

**Value iteration ($POMDP$, $\epsilon$)**
    **initialize** $V$ for all $b \in \mathcal{I}$;
    **repeat**
          $V' \leftarrow V$;
          update $V \leftarrow HV'$ for all $b \in \mathcal{I}$;
    **until** $\sup_b \mid V(b) - V'(b) \mid \leq \epsilon$
    **return** $V$;

Figure 3: Value iteration procedure.

converges to the unique fixed-point solution which is the direct consequence of Banach's theorem for contraction mappings (see, for example, Puterman (1994)).

In practice, we stop the iteration well before it reaches the limit solution. The stopping criterion we use in our algorithm (Figure 3) examines the maximum difference between value functions obtained in two consecutive steps — the so-called Bellman error (Puterman, 1994; Littman, 1996). The algorithm stops when this quantity falls below the threshold $\epsilon$. The accuracy of the approximate solution ($i$th value function) with regard to $V^*$ can be expressed in terms of the Bellman error $\epsilon$.

**Theorem 1** *Let $\epsilon = \sup_b |V_i(b) - V_{i-1}(b)| = \|V_i - V_{i-1}\|$ be the magnitude of the Bellman error. Then $\|V_i - V^*\| \leq \frac{\gamma\epsilon}{1-\gamma}$ and $\|V_{i-1} - V^*\| \leq \frac{\epsilon}{1-\gamma}$ hold.*

Then, to obtain the approximation of $V^*$ with precision $\delta$ the Bellman error should fall below $\frac{\delta(1-\gamma)}{\gamma}$.

### 2.4.1 PIECEWISE LINEAR AND CONVEX APPROXIMATIONS OF THE VALUE FUNCTION

The major difficulty in applying the value iteration (or dynamic programming) to belief-state MDPs is that the belief space is infinite and we need to compute an update $V_i = HV_{i-1}$ for all of it. This poses the following threats: the value function for the $i$th step may not be representable by finite means and/or computable in a finite number of steps.

To address this problem Sondik (Sondik, 1971; Smallwood & Sondik, 1973) showed that one can guarantee the computability of the $i$th value function as well as its finite description for a belief-state MDP by considering only piecewise linear and convex representations of value function estimates (see Figure 4). In particular, Sondik showed that for a piecewise linear and convex representation of $V_{i-1}$, $V_i = HV_{i-1}$ is computable and remains piecewise linear and convex.

**Theorem 2** *(Piecewise linear and convex functions). Let $V_0$ be an initial value function that is piecewise linear and convex. Then the $i$th value function obtained after a finite number of update steps for a belief-state MDP is also finite, piecewise linear and convex, and is equal to:*

$$V_i(b) = \max_{\alpha_i \in \Gamma_i} \sum_{s \in S} b(s)\alpha_i(s),$$

*where $b$ and $\alpha_i$ are vectors of size $|S|$ and $\Gamma_i$ is a finite set of vectors (linear functions) $\alpha_i$.*

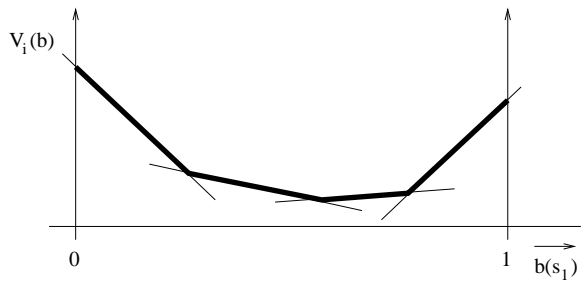

Figure 4: A piecewise linear and convex function for a POMDP with two process states $\{s_1, s_2\}$. Note that $b(s_1) = 1 - b(s_2)$ holds for any belief state.

The key part of the proof is that we can express the update for the $i$th value function in terms of linear functions $\Gamma_{i-1}$ defining $V_{i-1}$:

$$V_i(b) = \max_{a \in A} \left\{ \sum_{s \in S} \rho(s, a) b(s) + \gamma \sum_{o \in \Theta} \max_{\alpha_{i-1} \in \Gamma_{i-1}} \sum_{s' \in S} \left[ \sum_{s \in S} P(s', o|s, a) b(s) \right] \alpha_{i-1}(s') \right\}. \quad (6)$$

This leads to a piecewise linear and convex value function $V_i$ that can be represented by a finite set of linear functions $\alpha_i$, one linear function for every combination of actions and permutations of $\alpha_{i-1}$ vectors of size $|\Theta|$. Let $W = (a, \{o_1, \alpha_{i-1}^{j_1}\}, \{o_2, \alpha_{i-1}^{j_2}\}, \cdots \{o_{|\Theta|}, \alpha_{i-1}^{j_{|\Theta|}}\})$ be such a combination. Then the linear function corresponding to it is defined as

$$\alpha_i^W(s) = \rho(s, a) + \gamma \sum_{o \in \Theta} \sum_{s' \in S} P(s', o|s, a) \alpha_{i-1}^{j_o}(s'). \quad (7)$$

Theorem 2 is the basis of the dynamic programming algorithm for finding the optimal solution for the finite-horizon models and the value-iteration algorithm for finding near-optimal approximations of $V^*$ for the discounted, infinite-horizon model. Note, however, that this result does not imply piecewise linearity of the optimal (fixed-point) solution $V^*$.

### 2.4.2 ALGORITHMS FOR COMPUTING VALUE-FUNCTION UPDATES

The key part of the value-iteration algorithm is the computation of value-function updates $V_i = HV_{i-1}$. Assume an $i$th value function $V_i$ that is represented by a finite number of linear segments ($\alpha$ vectors). The total number of all its possible linear functions is $|A||\Gamma_{i-1}|^{|\Theta|}$ (one for every combination of actions and permutations of $\alpha_{i-1}$ vectors of size $|\Theta|$) and they can be enumerated in $O(|A||S|^2|\Gamma_{i-1}|^{|\Theta|})$ time. However, the complete set of linear functions is rarely needed: some of the linear functions are dominated by others and their omission does not change the resulting piecewise linear and convex function. This is illustrated in Figure 5.

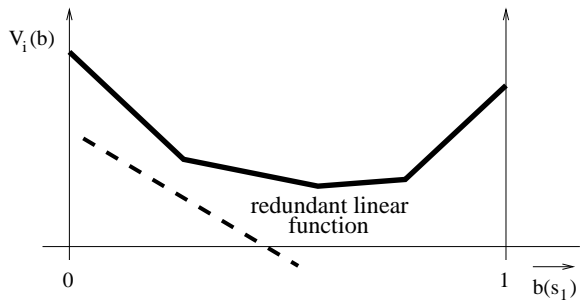

Figure 5: Redundant linear function. The function does not dominate in any of the regions of the belief space and can be excluded.

A linear function that can be eliminated without changing the resulting value function solution is called *redundant*. Conversely, a linear function that singlehandedly achieves the optimal value for at least one point of the belief space is called *useful*.[4]

For the sake of computational efficiency it is important to make the size of the linear function set as small as possible (keep only useful linear functions) over value-iteration steps. There are two main approaches for computing useful linear functions. The first approach is based on a generate-and-test paradigm and is due to Sondik (1971) and Monahan (1982). The idea here is to enumerate all possible linear functions first, then test the usefulness of linear functions in the set and prune all redundant vectors. Recent extensions of the method interleave the generate and test stages and do early pruning on a set of partially constructed linear functions (Zhang & Liu, 1997a; Cassandra, Littman, & Zhang, 1997; Zhang & Lee, 1998).

The second approach builds on Sondik's idea of computing a useful linear function for a single belief state (Sondik, 1971; Smallwood & Sondik, 1973), which can be done efficiently. The key problem here is to locate all belief points that seed useful linear functions and different methods address this problem differently. Methods that implement this idea are Sondik's one- and two-pass algorithms (Sondik, 1971), Cheng's methods (Cheng, 1988), and the Witness algorithm (Kaelbling, Littman, & Cassandra, 1999; Littman, 1996; Cassandra, 1998).

### 2.4.3 LIMITATIONS AND COMPLEXITY

The major difficulty in solving a belief-state MDP is that the complexity of a piecewise linear and convex function can grow extremely fast with the number of update steps. More specifically, the size of a linear function set defining the function can grow exponentially (in the number of observations) during a single update step. Then, assuming that the initial value function is linear, the number of linear functions defining the $i$th value function is $O(|A|^{|\Theta|^{i-1}})$.

---

4. In defining redundant and useful linear functions we assume that there are no linear function duplicates, i.e. only one copy of the same linear function is kept in the set $\Gamma_i$.

The potential growth of the size of the linear function set is not the only bad news. As remarked earlier, a piecewise linear convex value function is usually less complex than the worst case because many linear functions can be pruned away during updates. However, it turned out that the task of identifying all useful linear functions is computationally intractable as well (Littman, 1996). This means that one faces not only the potential super-exponential growth of the number of useful linear functions, but also inefficiencies related to the identification of such vectors. This is a significant drawback that makes the exact methods applicable only to relatively simple problems.

The above analysis suggests that solving a POMDP problem is an intrinsically hard task. Indeed, finding the optimal solution for the finite-horizon problem is PSPACE-hard (Papadimitriou & Tsitsiklis, 1987). Finding the optimal solution for the discounted infinite-horizon criterion is even harder. The corresponding decision problem has been shown to be undecidable (Madani et al., 1999), and thus the optimal solution may not be computable.

### 2.4.4 STRUCTURAL REFINEMENTS OF THE BASIC ALGORITHM

The standard POMDP model uses a flat state space and full transition and reward matrices. However, in practice, problems often exhibit more structure and can be represented more compactly, for example, using graphical models (Pearl, 1988; Lauritzen, 1996), most often dynamic belief networks (Dean & Kanazawa, 1989; Kjaerulff, 1992) or dynamic influence diagrams (Howard & Matheson, 1984; Tatman & Schachter, 1990).[5] There are many ways to take advantage of the problem structure to modify and improve exact algorithms. For example, a refinement of the basic Monahan algorithm to compact transition and reward models has been studied by Boutilier and Poole (1996). A hybrid framework that combines MDP-POMDP problem-solving techniques to take advantage of perfectly and partially observable components of the model and the subsequent value function decomposition was proposed by Hauskrecht (1997, 1998, 2000). A similar approach with perfect information about a region (subset of states) containing the actual underlying state was discussed by Zhang and Liu (1997b, 1997a). Finally, Castañon (1997) and Yost (1998) explore techniques for solving large POMDPs that consist of a set of smaller, resource-coupled but otherwise independent POMDPs.

## 2.5 Extracting Control Strategy

Value iteration allow us to compute an $i$th approximation of the value function $V_i$. However, our ulimate goal is to find the optimal control strategy $\mu^* : \mathcal{I} \to A$ or its close approximation. Thus our focus here is on the problem of extraction of control strategies from the results of value iteration.

### 2.5.1 LOOKAHEAD DESIGN

The simplest way to define the control function $\mu : \mathcal{I} \to A$ from the value function $V_i$ is via greedy one-step lookahead:

$$\mu(b) = \arg\max_{a \in A} \left\{ \sum_{s \in S} \rho(s, a) b(s) + \gamma \sum_{o \in \Theta} P(o|b, a) V_i(\tau(b, o, a)) \right\}.$$

---

5. See the survey by Boutilier, Dean and Hanks (1999) for different ways to represent structured MDPs.

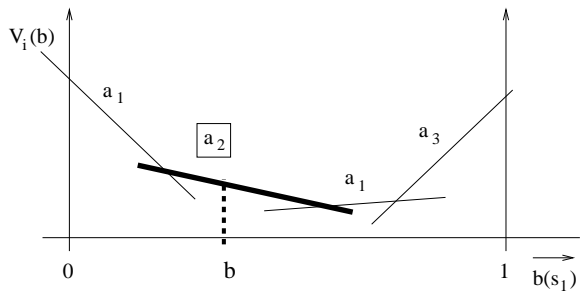

Figure 6: Direct control design. Every linear function defining $V_i$ is associated with an action. The action is selected if its linear function (or Q-function) is maximal.

As $V_i$ represents only the $i$th approximation of the optimal value function, the question arises how good the resulting controller really is.[6] The following theorem (Puterman, 1994; Williams & Baird, 1994; Littman, 1996) relates the accuracy of the (lookahead) controller and the Bellman error.

**Theorem 3** *Let $\epsilon = \|V_i - V_{i-1}\|$ be the magnitude of the Bellman error. Let $V_i^{LA}$ be the expected reward for the lookahead controller designed for $V_i$. Then $\|V_i^{LA} - V^*\| \leq \frac{2\epsilon\gamma}{1-\gamma}$.*

The bound can be used to construct the value-iteration routine that yields a lookahead strategy with a minimum required precision. The result can be also extended to the $k$-step lookahead design in a straightforward way; with $k$ steps, the error bound becomes $\|V_i^{LA(k)} - V^*\| \leq \frac{2\epsilon\gamma^k}{(1-\gamma)}$.

### 2.5.2 DIRECT DESIGN

To extract the control action via lookahead essentially requires computing one full update. Obviously, this can lead to unwanted delays in reaction times. In general, we can speed up the response by remembering and using additional information. In particular, every linear function defining $V_i$ is associated with the choice of action (see Equation 7). The action is a byproduct of methods for computing linear functions and no extra computation is required to find it. Then the action corresponding to the best linear function can be selected directly for any belief state. The idea is illustrated in Figure 6.

The bound on the accuracy of the direct controller for the infinite-horizon case can be once again derived in terms of the magnitude of the Bellman error.

**Theorem 4** *Let $\epsilon = \|V_i - V_{i-1}\|$ be the magnitude of the Bellman error. Let $V_i^{DR}$ be an expected reward for the direct controller designed for $V_i$. Then $\|V_i^{DR} - V^*\| \leq \frac{2\epsilon}{1-\gamma}$.*

The direct action choice is closely related to the notion of action-value function (or Q-function). Analogously to Equation 4, the $i$th Q-function satisfies

$$V_i(b) = \max_{a \in A} Q_i(b, a),$$

---

6. Note that the control action extracted via lookahead from $V_i$ is optimal for $(i + 1)$ steps-to-go and the finite-horizon model. The main difference here is that $V_i$ is the optimal value function for $i$ steps to go.

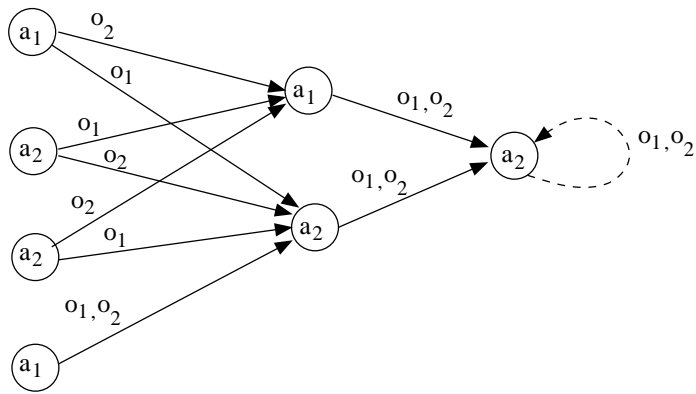

Figure 7: A policy graph (finite-state machine) obtained after two value iteration steps. Nodes correspond to linear functions (or states of the finite-state machine) and links to dependencies between linear functions (transitions between states). Every linear function (node) is associated with an action. To ensure that the policy can be also applied to the infinite-horizon problem, we add a cycle to the last state (dashed line).

$$Q_i(b, a) = R(b, a) + \gamma \sum_{o \in \Theta} P(o|b, a) V_{i-1}(\tau(b, a, o)).$$

From this perspective, the direct strategy selects the action with the best (maximum) Q-function for a given belief state.[7]

### 2.5.3 FINITE-STATE MACHINE DESIGN

A more complex refinement of the above technique is to remember, for every linear function in $V_i$, not only the action choice but also the choice of a linear function for the previous step and to do this for all observations (see Equation 7). As the same idea can be applied recursively to the linear functions for all previous steps, we can obtain a relatively complex dependency structure relating linear functions in $V_i, V_{i-1}, \cdots V_0$, observations and actions that itself represents a control strategy (Kaelbling et al., 1999).

To see this, we model the structure in graphical terms (Figure 7). Here different nodes represent linear functions, actions associated with nodes correspond to optimizing actions, links emanating from nodes correspond to different observations, and successor nodes correspond to linear functions paired with observations. Such graphs are also called *policy graphs* (Kaelbling et al., 1999; Littman, 1996; Cassandra, 1998). One interpretation of the dependency structure is that it represents a collection of finite-state machines (FSMs) with many possible initial states that implement a POMDP controller: nodes correspond to states of the controller, actions to controls (outputs), and links to transitions conditioned on inputs (observations). The start state of the FSM controller is chosen greedily by selecting the linear function (controller state) optimizing the value of an initial belief state.

---

7. Williams and Baird (1994) also give results relating the accuracy of the direct Q-function controller to the Bellman error of Q-functions.

The advantage of the finite-state machine representation of the strategy is that for the first $i$ steps it works with observations directly; belief-state updates are not needed. This contrasts with the other two policy models (lookahead and direct models), which must keep track of the current belief state and update it over time in order to extract appropriate control. The drawback of the approach is that the FSM controller is limited to $i$ steps that correspond to the number of value iteration steps performed. However, in the infinite-horizon model the controller is expected to run for an infinite number of steps. One way to remedy this deficiency is to extend the FSM structure and to create cycles that let us visit controller states repeatedly. For example, adding a cycle transition to the end state of the FSM controller in Figure 7 (dashed line) ensures that the controller is also applicable to the infinite-horizon problem.

## 2.6 Policy Iteration

An alternative method for finding the solution for the discounted infinite-horizon problem is *policy iteration* (Howard, 1960; Sondik, 1978). Policy iteration searches the policy space and gradually improves the current control policy for one or more belief states. The method consists of two steps performed iteratively:

- *policy evaluation*: computes expected value for the current policy;

- *policy improvement*: improves the current policy.

As we saw in Section 2.5, there are many ways to represent a control policy for a POMDP. Here we restrict attention to a finite-state machine model in which observations correspond to inputs and actions to outputs (Platzman, 1980; Hansen, 1998b; Kaelbling et al., 1999).[8]

### 2.6.1 Finite-State Machine Controller

A finite-state machine (FSM) controller $C = (M, \Theta, A, \phi, \eta, \psi)$ for a POMDP is described by a set of memory states $M$ of the controller, a set of observations (inputs) $\Theta$, a set of actions (outputs) $A$, a transition function $\phi : M \times \Theta \to M$ mapping states of the FSM to next memory states given the observation, and an output function $\eta : M \to A$ mapping memory states to actions. A function $\psi : \mathcal{I}_0 \to M$ selects the initial memory state given the initial information state. The initial information state corresponds either to a prior or a posterior belief state at time $t_0$ depending on the availability of an initial observation.

### 2.6.2 Policy Evaluation

The first step of the policy iteration is policy evaluation. The most important property of the FSM model is that the value function for a specific FSM strategy can be computed efficiently in the number of controller states $M$. The key to efficient computability is the fact that the value function for executing an FSM strategy from some memory state $x$ is linear (Platzman, 1980).[9]

---

8. A policy-iteration algorithm in which policies are defined over the regions of the belief space was described first by Sondik (1978).

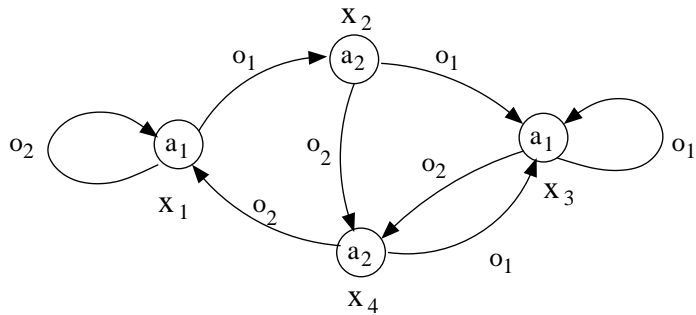

Figure 8: An example of a four-state FSM policy. Nodes represent states, links transitions between states (conditioned on observations). Every memory state has an associated control action (output).

**Theorem 5** *Let $C$ be a finite-state machine controller with a set of memory states $M$. The value function for applying $C$ from a memory state $x \in M$, $V^C(x, b)$, is linear. Value functions for all $x \in M$ can be found by solving a system of linear equations with $|S||M|$ variables.*

We illustrate the main idea by an example. Assume an FSM controller with four memory states $\{x_1, x_2, x_3, x_4\}$, as in Figure 8, and a stochastic process with two hidden states $S = \{s_1, s_2\}$. The value of the policy for an augmented state space $S \times M$ satisfies a system of linear equations

$$V(x_1, s_1) = \rho(s_1, \eta(x_1)) + \gamma \sum_{o \in \Theta} \sum_{s \in S} P(o, s | s_1, \eta(x_1)) V(\phi(x_1, o), s)$$

$$V(x_1, s_2) = \rho(s_2, \eta(x_1)) + \gamma \sum_{o \in \Theta} \sum_{s \in S} P(o, s | s_2, \eta(x_1)) V(\phi(x_1, o), s)$$

$$V(x_2, s_1) = \rho(s_1, \eta(x_2)) + \gamma \sum_{o \in \Theta} \sum_{s \in S} P(o, s | s_1, \eta(x_2)) V(\phi(x_2, o), s)$$

$$\dots$$

$$V(x_4, s_2) = \rho(s_2, \eta(x_4)) + \gamma \sum_{o \in \Theta} \sum_{s \in S} P(o, s | s_2, \eta(x_4)) V(\phi(x_4, o), s),$$

where $\eta(x)$ is the action executed in $x$ and $\phi(x, o)$ is the state to which one transits after seeing an input (observation) $o$. Assuming we start the policy from the memory state $x_1$, the value of the policy is:

$$V^C(x_1, b) = \sum_{s \in S} V(x_1, s) b(s).$$

---

9. The idea of linearity and efficient computability of the value functions for a fixed FSM-based strategy has been addressed recently in different contexts by a number of researchers (Littman, 1996; Cassandra, 1998; Hauskrecht, 1997; Hansen, 1998b; Kaelbling et al., 1999). However, the origins of the idea can be traced to the earlier work by Platzman (1980).

Thus the value function is linear and can be computed efficiently by solving a system of linear equations.

Since in general the FSM controller can start from any memory state, we can always choose the initial memory state greedily, maximizing the expected value of the result. In such a case the optimal choice function $\psi$ is defined as:

$$\psi(b) = \arg\max_{x \in M} V^C(x, b),$$

and the value for the FSM policy C and belief state $b$ is:

$$V^C(b) = \max_{x \in M} V^C(x, b) = V^C(\psi(b), b).$$

Note that the resulting value function for the strategy $C$ is piecewise linear and convex and represents expected rewards for following $C$. Since no strategy can perform better that the optimal strategy, $V^C \leq V^*$ must hold.

### 2.6.3 POLICY IMPROVEMENT

The policy-iteration method, searching the space of controllers, starts from an arbitrary initial policy and improves it gradually by refining its finite-state machine (FSM) description. In particular, one keeps modifying the structure of the controller by adding or removing controller states (memory) and transitions. Let $C$ and $C'$ be an old and a new FSM controller. In the improvement step we must satisfy

$$V^{C'}(b) \geq V^C(b) \text{ for all } b \in \mathcal{I};$$

$$\exists b \in \mathcal{I} \text{ such that } V^{C'}(b) > V^C(b).$$

To guarantee the improvement, Hansen (1998a, 1998b) proposed a policy-iteration algorithm that relies on exact value function updates to obtain a new improved policy structure.[10] The basic idea of the improvement is based on the observation that one can switch back and forth between the FSM policy description and the piecewise-linear and convex representation of a value function. In particular:

- the value function for an FSM policy is piecewise-linear and convex and every linear function describing it corresponds to a memory state of a controller;

- individual linear functions comprising the new value function after an update can be viewed as new memory states of an FSM policy, as described in Section 2.5.3.

This allows us to improve the policy by adding new memory states corresponding to linear functions of the new value function obtained after the exact update. The technique can be refined by removing some of the linear functions (memory states) whenever they are fully dominated by one of the other linear functions.

---

10. A policy-iteration algorithm that exploits exact value function updates but works with policies defined over the belief space was used earlier by Sondik (1978).

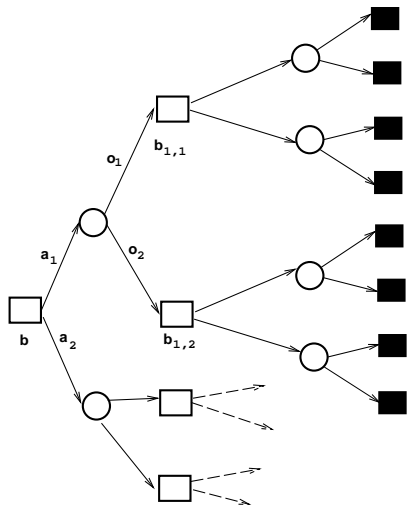

Figure 9: A two-step decision tree. Rectangles correspond to the decision nodes (moves of the decision-maker) and circles to chance nodes (moves of the environment). Black rectangles represent leaves of the tree. The reward for a specific path is associated with every leaf of the tree. Decision nodes are associated with information states obtained by following action and observation choices along the path from the root of the tree. For example, $b_{1,1}$ is a belief state obtained by performing action $a_1$ from the initial belief state $b$ and observing observation $o_1$.

## 2.7 Forward (Decision Tree) Methods

The methods discussed so far assume no prior knowledge of the initial belief state and treat all belief states as equally likely. However, if the initial state is known and fixed, methods can often be modified to take advantage of this fact. For example, for the finite-horizon problem, only a finite number of belief states can be reached from a given initial state. In this case it is very often easier to enumerate all possible histories (sequences of actions and observations) and represent the problem using stochastic decision trees (Raiffa, 1970). An example of a two-step decision tree is shown in Figure 9.

The algorithm for solving the stochastic decision tree basically mimics value-function updates, but is restricted only to situations that can be reached from the initial belief state. The key difficulty here is that the number of all possible trajectories grows exponentially with the horizon of interest.

### 2.7.1 COMBINING DYNAMIC-PROGRAMMING AND DECISION-TREE TECHNIQUES

To solve a POMDP for a fixed initial belief state, we can apply two strategies: one constructs the decision tree first and then solves it, the other solves the problem in a backward fashion via dynamic programming. Unfortunately, both these techniques are inefficient, one suffering from exponential growth in the decision tree size, the other from super-exponential growth in the value function complexity. However, the two techniques can be combined in a way that at least partially eliminates their disadvantages. The idea is based on the fact that the two techniques work on the solution from two different sides (one forward and the other backward) and the complexity for each of them worsens gradually. Then the solution is to compute the complete $k$th value function using dynamic programming (value iteration) and cover the remaining steps by forward decision-tree expansion.

Various modifications of the above idea are possible. For example, one can often replace exact dynamic programming with two more efficient approximations providing upper and lower bounds of the value function. Then the decision tree must be expanded only when the bounds are not sufficient to determine the optimal action choice. A number of search techniques developed in the AI literature (Korf, 1985) combined with branch-and-bound pruning (Satia & Lave, 1973) can be applied to this type of problem. Several researchers have experimented with them to solve POMDPs (Washington, 1996; Hauskrecht, 1997; Hansen, 1998b). Other methods applicable to this problem are based on Monte-Carlo sampling (Kearns, Mansour, & Ng, 1999; McAllester & Singh, 1999) and real-time dynamic programming (Barto, Bradtke, & Singh, 1995; Dearden & Boutilier, 1997; Bonet & Geffner, 1998).

### 2.7.2 CLASSICAL PLANNING FRAMEWORK

POMDP problems with fixed initial belief states and their solutions are closely related to work in classical planning and its extensions to handle stochastic and partially observable domains, particularly the work on BURIDAN and C-BURIDAN planners (Kushmerick, Hanks, & Weld, 1995; Draper, Hanks, & Weld, 1994). The objective of these planners is to maximize the probability of reaching some goal state. However, this task is similar to the discounted reward task in terms of complexity, since a discounted reward model can be converted into a goal-achievement model by introducing an absorbing state (Condon, 1992).

## 3. Heuristic Approximations

The key obstacle to wider application of the POMDP framework is the computational complexity of POMDP problems. In particular, finding the optimal solution for the finite-horizon case is PSPACE-hard (Papadimitriou & Tsitsiklis, 1987) and the discounted infinite-horizon case may not even be computable (Madani et al., 1999). One approach to such problems is to approximate the solution to some $\epsilon$-precision. Unfortunately, even this remains intractable and in general POMDPs cannot be approximated efficiently (Burago, Rougemont, & Slissenko, 1996; Lusena, Goldsmith, & Mundhenk, 1998; Madani et al., 1999). This is also the reason why only very simple problems can be solved optimally or near-optimally in practice.

To alleviate the complexity problem, research in the POMDP area has focused on various heuristic methods (or approximations without the error parameter) that are more efficient.[11] Heuristic methods are also our focus here. Thus, when referring to approximations, we mean heuristics, unless specifically stated otherwise.

---

11. The quality of a heuristic approximation can be tested using the Bellman error, which requires one exact update step. However, heuristic methods per se do not contain a precision parameter.

The many approximation methods and their combinations can be divided into two often very closely related classes: *value-function approximations* and *policy approximations*.

## 3.1 Value-Function Approximations

The main idea of the value-function approximation approach is to approximate the optimal value function $V : \mathcal{I} \to \rm I\!R$ with a function $\widehat{V} : \mathcal{I} \to \rm I\!R$ defined over the same information space. Typically, the new function is of lower complexity (recall that the optimal or near-optimal value function may consist of a large set of linear functions) and is easier to compute than the exact solution. Approximations can be often formulated as dynamic programming problems and can be expressed in terms of approximate value-function updates $\widehat{H}$. Thus, to understand the differences and advantages of various approximations and exact methods, it is often sufficient to analyze and compare their update rules.

### 3.1.1 VALUE-FUNCTION BOUNDS

Although heuristic approximations have no guaranteed precision, in many cases we are able to say whether they overestimate or underestimate the optimal value function. The information on bounds can be used in multiple ways. For example, upper- and lower-bounds can help in narrowing the range of the optimal value function, elimination of some of the suboptimal actions and subsequent speed-ups of exact methods. Alternatively, one can use knowledge of both value-function bounds to determine the accuracy of a controller generated based on one of the bounds (see Section 3.1.3). Also, in some instances, a lower bound alone is sufficient to guarantee the control choice that always achieves an expected reward at least as high as the one given by that bound (Section 4.7.2).

The bound property of different methods can be determined by examining the updates and their bound relations.

**Definition 5** (*Upper bound*). *Let $H$ be the exact value-function mapping and $\widehat{H}$ its approximation. We say that $\widehat{H}$ upper-bounds $H$ for some $V$ when $(\widehat{H}V)(b) \geq (HV)(b)$ holds for every $b \in \mathcal{I}$.*

An analogous definition can be constructed for the lower bound.

### 3.1.2 CONVERGENCE OF APPROXIMATE VALUE ITERATION

Let $\widehat{H}$ be a value-function mapping representing an approximate update. Then the approximate value iteration computes the $i$th value function as $\widehat{V}_i = \widehat{H}\widehat{V}_{i-1}$. The fixed-point solution $\widehat{V}^* = \widehat{H}\widehat{V}^*$ or its close approximation would then represent the intended output of the approximation routine. The main problem with the iteration method is that in general it can converge to unique or multiple solutions, diverge, or oscillate, depending on $\widehat{H}$ and the initial function $\widehat{V}_0$. Therefore, unique convergence cannot be guaranteed for an arbitrary mapping $\widehat{H}$ and the convergence of a specific approximation method must be proved.

**Definition 6** (*Convergence of $\widehat{H}$*). *The value iteration with $\widehat{H}$ converges for a value function $V_0$ when $\lim_{n \to \infty} (\widehat{H}^n V_0)$ exists.*

**Definition 7** (*Unique convergence of $\widehat{H}$*). *The value iteration converges uniquely for $\mathcal{V}$ when for every $V \in \mathcal{V}$, $\lim_{n \to \infty}(\widehat{H}^n V)$ exists and for all pairs $V, U \in \mathcal{V}$, $\lim_{n \to \infty}(\widehat{H}^n V) = \lim_{n \to \infty}(\widehat{H}^n U)$.*

A sufficient condition for the unique convergence is to show that $\widehat{H}$ be a contraction. The contraction and the bound properties of $\widehat{H}$ can be combined, under additional conditions, to show the convergence of the iterative approximation method to the bound. To address this issue we present a theorem comparing fixed-point solutions of two value-function mappings.

**Theorem 6** *Let $H_1$ and $H_2$ be two value-function mappings defined on $\mathcal{V}_1$ and $\mathcal{V}_2$ such that*

1. *$H_1$, $H_2$ are contractions with fixed points $V_1^*$, $V_2^*$;*

2. *$V_1^* \in \mathcal{V}_2$ and $H_2 V_1^* \geq H_1 V_1^* = V_1^*$;*

3. *$H_2$ is an isotone mapping.*

*Then $V_2^* \geq V_1^*$ holds.*

Note that this theorem does not require that $\mathcal{V}_1$ and $\mathcal{V}_2$ cover the same space of value functions. For example, $\mathcal{V}_2$ can cover all possible value functions of a belief-state MDP, while $\mathcal{V}_1$ can be restricted to a space of piecewise linear and convex value functions. This gives us some flexibility in the design of iterative approximation algorithms for computing value-function bounds. An analogous theorem also holds for the lower bound.

### 3.1.3 CONTROL

Once the approximation of the value-function is available, it can be used to generate a control strategy. In general, control solutions correspond to options presented in Section 2.5 and include lookahead, direct (Q-function) and finite-state machine designs.

A drawback of control strategies based on heuristic approximations is that they have no precision guarantee. One way to find the accuracy of such strategies is to do one exact update of the value function approximation and adopt the result of Theorems 1 and 3 for the Bellman error. An alternative solution to this problem is to bound the accuracy of such controllers using the upper- and the lower-bound approximations of the optimal value function. To illustrate this approach, we present and prove (in the Appendix) the following theorem that relates the quality of bounds to the quality of a lookahead controller.

**Theorem 7** *Let $\widehat{V}_U$ and $\widehat{V}_L$ be upper and lower bounds of the optimal value function for the discounted infinite-horizon problem. Let $\epsilon = \sup_b |\widehat{V}_U(b) - \widehat{V}_L(b)| = \|\widehat{V}_U - \widehat{V}_L\|$ be the maximum bound difference. Then the expected reward for a lookahead controller $\widehat{V}^{LA}$, constructed for either $\widehat{V}_U$ or $\widehat{V}_L$, satisfies $\|\widehat{V}^{LA} - V^*\| \leq \frac{\epsilon(2-\gamma)}{(1-\gamma)}$.*

## 3.2 Policy Approximation

An alternative to value-function approximation is policy approximation. As shown earlier, a strategy (controller) for a POMDP can be represented using a finite-state machine (FSM) model. The policy iteration searches the space of all possible policies (FSMs) for the optimal or near-optimal solution. This space is usually enormous, which is the bottleneck of the method. Thus, instead of searching the complete policy space, we can restrict our attention only to its subspace that we believe to contain the optimal solution or a good approximation. Memoryless policies (Platzman, 1977; White & Scherer, 1994; Littman, 1994; Singh, Jaakkola, & Jordan, 1994), policies based on truncated histories (Platzman, 1977; White & Scherer, 1994; McCallum, 1995), or finite-state controllers with a fixed number of memory states (Platzman, 1980; Hauskrecht, 1997; Hansen, 1998a, 1998b) are all examples of a policy-space restriction. In the following we consider only the finite-state machine model (see Section 2.6.1), which is quite general; other models can be viewed as its special cases.

States of an FSM policy model represent the memory of the controller and, in general, summarize information about past activities and observations. Thus, they are best viewed as approximations of the information states, or as *feature states*. The transition model of the controller ($\phi$) then approximates the update function of the information-state MDP ($\tau$) and the output function of an FSM ($\eta$) approximates the control function ($\mu$) mapping information states to actions. The important property of the model, as shown Section 2.6.2, is that the value function for a fixed controller and fixed initial memory state can be obtained efficiently by solving a system of linear equations (Platzman, 1980).

To apply the policy approximation approach we first need to decide (1) how to restrict a space of policies and (2) how to judge the policy quality.

A restriction frequently used is to consider only controllers with a fixed number of states, say $k$. Other structural restrictions further narrowing the space of policies can restrict either the output function (choice of actions at different controller states), or the transitions between the current and next states. In general, any heuristic or domain-related insight may help in selecting the right biases.

Two different policies can yield value functions that are better in different regions of the belief space. Thus, in order to decide which policy is the best, we need to define the importance of different regions and their combinations. There are multiple solutions to this. For example, Platzman (1980) considers the worst-case measure and optimizes the worst (minimal) value for all initial belief states. Let $\mathcal{C}$ be a space of FSM controllers satisfying given restrictions. Then the quality of a policy under the worst case measure is:

$$\max_{C \in \mathcal{C}} \min_{b \in \mathcal{I}} \max_{x \in M_C} V^C(x, b).$$

Another option is to consider a distribution over all initial belief states and maximize the expectation of their value function values. However, the most common objective is to choose the policy that leads to the best value for a single initial belief state $b_0$:

$$\max_{C \in \mathcal{C}} \max_{x \in M_C} V^C(x, b_0).$$

Finding the optimal policy for this case reduces to a combinatorial optimization problem. Unfortunately, for all but trivial cases, even this problem is computationally intractable. For example, the problem of finding the optimal policy for a memoryless case (only current observations are considered) is NP-hard (Littman, 1994). Thus, various heuristics are typically applied to alleviate this difficulty (Littman, 1994).

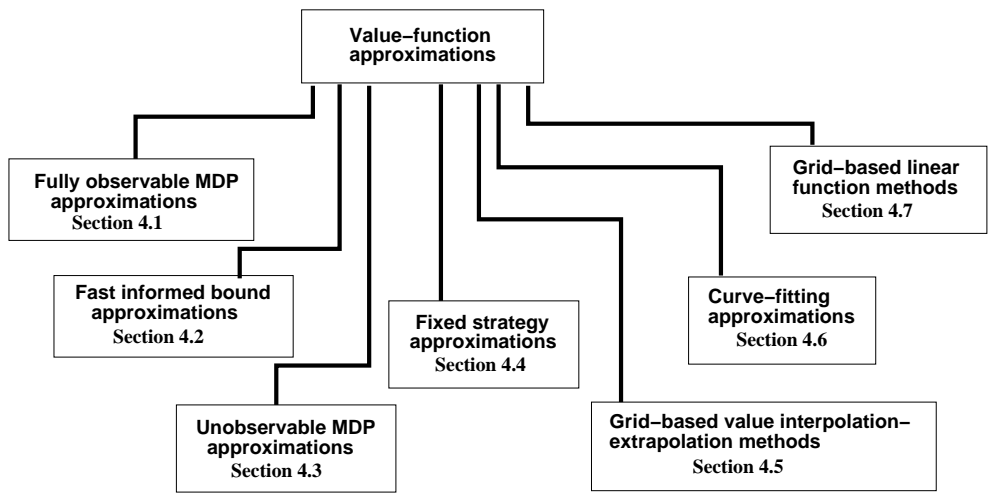

Figure 10: Value-function approximation methods.

### 3.2.1 RANDOMIZED POLICIES

By restricting the space of policies we simplify the policy optimization problem. On the other hand, we simultaneously give up an opportunity to find the best optimal policy, replacing it with the best restricted policy. Up to this point, we have considered only deterministic policies with a fixed number of internal controller states, that is, policies with deterministic output and transition functions. However, finding the best deterministic policy is not always the best option: *randomized policies*, with randomized output and transition functions, usually lead to the far better performance. The application of randomized (or stochastic) policies to POMDPs was introduced by Platzman (1980). Essentially, any deterministic policy can be represented as a randomized policy with a single action and transition, so that the best randomized policy is no worse than the best deterministic policy. The difference in control performance of two policies shows up most often in cases when the number of states of the controller is relatively small compared to that in the optimal strategy.

The advantage of stochastic policies is that their space is larger and parameters of the policy are continuous. Therefore the problem of finding the optimal stochastic policy becomes a non-linear optimization problem and a variety of optimization methods can be applied to solve it. An example is the gradient-based approach (see Meuleau et al., 1999).

## 4. Value-Function Approximation Methods

In this section we discuss in more depth value-function approximation methods. We focus on approximations with belief information space.[12] We survey known techniques, but also include a number of new methods and modifications of existing methods. Figure 10 summarizes the methods covered. We describe the methods by means of update rules they implement, which simplifies their analysis and theoretical comparison. We focus on the following properties: the complexity of the dynamic-programming (value-iteration) updates; the complexity of value functions each method uses; the ability of methods to bound the exact update; the convergence of value iteration with approximate update rules; and the control performance of related controllers. The results of the theoretical analysis are illustrated empirically on a problem from the agent-navigation domain. In addition, we use the agent navigation problem to illustrate and give some intuitions on other characteristics of methods with no theoretical underpinning. Thus, these results should not be generalized to other problems or used to rank different methods.

---

12. Alternative value-function approximations may work with complete histories of past actions and observations. Approximation methods used by White and Scherer (1994) are an example.

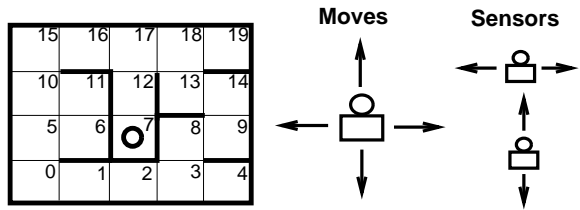

Figure 11: Test example. The maze navigation problem: Maze20.

### AGENT-NAVIGATION PROBLEM

Maze20 is a maze-navigation problem with 20 states, six actions and eight observations. The maze (Figure 11) consists of 20 partially connected rooms (states) in which a robot operates and collects rewards. The robot can move in four directions (north, south, east and west) and can check for the presence of walls using its sensors. But, neither "move" actions nor sensor inputs are perfect, so that the robot can end up moving in unintended directions. The robot moves in an unintended direction with probability of 0.3 (0.15 for each of the neighboring directions). A move into the wall keeps the robot in the same position. Investigative actions help the robot to navigate by activating sensor inputs. Two such investigative actions allow the robot to check inputs (presence of a wall) in the north-south and east-west directions. Sensor accuracy in detecting walls is 0.75 for a two-wall case (e.g. both north and south wall), 0.8 for a one-wall case (north or south) and 0.89 for a no-wall case, with smaller probabilities for wrong perceptions.

The control objective is to maximize the expected discounted rewards with a discount factor of 0.9. A small reward is given for every action not leading to bumping into the wall (4 points for a move and 2 points for an investigative action), and one large reward (150 points) is given for achieving the special target room (indicated by the circle in the figure) and recognizing it by performing one of the move actions. After doing that and collecting the reward, the robot is placed at random in a new start position.

Although the Maze20 problem is of only moderate complexity with regard to the size of state, action and observation spaces, its exact solution is beyond the reach of current exact methods. The exact methods tried on the problem include the Witness algorithm (Kaelbling et al., 1999), the incremental pruning algorithm (Cassandra et al., 1997)[13] and policy iteration with an FSM model (Hansen, 1998b). The main obstacle preventing these algorithms from obtaining the optimal or close-to-optimal solution was the complexity of the value function (the number of linear functions needed to describe it) and subsequent running times and memory problems.

---

13. Many thanks to Anthony Cassandra for running these algorithms.

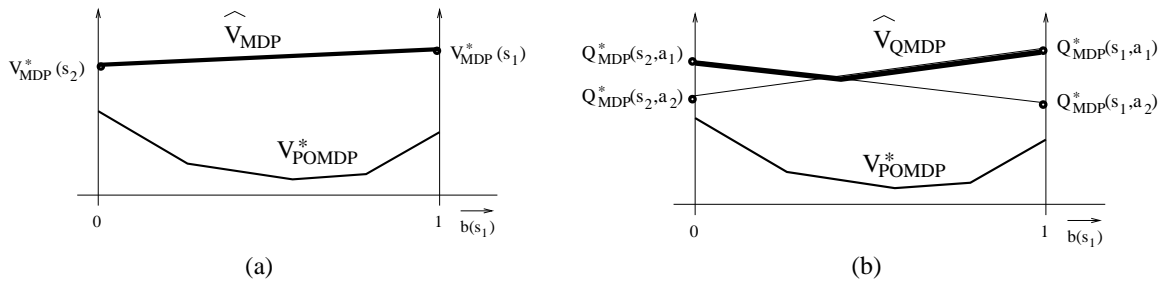

Figure 12: Approximations based on the fully observable version of a two state POMDP (with states $s_1, s_2$): (a) the MDP approximation; (b) the QMDP approximation. Values at extreme points of the belief space are solutions of the fully observable MDP.

## 4.1 Approximations with Fully Observable MDP

Perhaps the simplest way to approximate the value function for a POMDP is to assume that states of the process are fully observable (Astrom, 1965; Lovejoy, 1993). In that case the optimal value function $V^*$ for a POMDP can be approximated as:

$$\widehat{V}(b) = \sum_{s \in S} b(s) V^*_{MDP}(s), \tag{8}$$

where $V^*_{MDP}(s)$ is the optimal value function for state $s$ for the fully observable version of the process. We refer to this approximation as to the *MDP approximation*. The idea of the approximation is illustrated in Figure 12a. The resulting value function is linear and is fully defined by values at extreme points of the belief simplex. These correspond to the optimal values for the fully observable case. The main advantage of the approximation is that the fully observable MDP (FOMDP) can be solved efficiently for both the finite-horizon problem and discounted infinite-horizon problems.[14] The update step for the (fully observable) MDP is:

$$V^{MDP}_{i+1}(s) = \max_a \left\{ \rho(s, a) + \gamma \sum_{s' \in S} P(s'|s, a) V^{MDP}_i(s') \right\}.$$

---

14. The solution for the finite-state fully observable MDP and discounted infinite-horizon criterion can be found efficiently by formulating an equivalent linear programming task (Bertsekas, 1995)

### 4.1.1 MDP Approximation

The MDP-approximation approach (Equation 8) can be also described in terms of value-function updates for the belief-space MDP. Although this step is strictly speaking redundant here, it simplifies the analysis and comparison of this approach to other approximations.

Let $\widehat{V}_i$ be a linear value function described by a vector $\alpha_i^{MDP}$ corresponding to values of $V_i^{MDP}(s')$ for all states $s' \in S$. Then the $(i+1)$th value function $\widehat{V}_{i+1}$ is

$$\widehat{V}_{i+1}(b) = \sum_{s \in S} b(s) \max_{a \in A} \left[ \rho(s,a) + \gamma \sum_{s' \in S} P(s'|s,a)\alpha_i^{MDP}(s') \right]$$
$$= (H_{MDP}\widehat{V}_i)(b).$$

$\widehat{V}_{i+1}$ is described by a linear function with components

$$\alpha_{i+1}^{MDP}(s) = V_{i+1}^{MDP}(s) = \max_a \left\{ \rho(s,a) + \gamma \sum_{s \in S} P(s'|s,a)\alpha_i^{MDP}(s') \right\}.$$

The MDP-based rule $H_{MDP}$ can be also rewritten in a more general form that starts from an arbitrary piecewise linear and convex value function $V_i$, represented by a set of linear functions $\Gamma_i$:

$$\widehat{V}_{i+1}(b) = \sum_{s \in S} b(s) \max_{a \in A} \left\{ \rho(s,a) + \gamma \sum_{s' \in S} P(s'|s,a) \max_{\alpha_i \in \Gamma_i} \alpha_i(s') \right\}.$$

The application of the $H_{MDP}$ mapping always leads to a linear value function. The update is easy to compute and takes $O(|A||S|^2 + |\Gamma_i||S|)$ time. This reduces to $O(|A||S|^2)$ time when only MDP-based updates are strung together. As remarked earlier, the optimal solution for the infinite-horizon, discounted problem can be solved efficiently via linear programming.

The update for the MDP approximation upper-bounds the exact update, that is, $H\widehat{V}_i \leq H_{MDP}\widehat{V}_i$. We show this property later in Theorem 9, which covers more cases. The intuition is that we cannot get a better solution with less information, and thus the fully observable MDP must upper-bound the partially observable case.

### 4.1.2 Approximation with Q-Functions (QMDP)

A variant of the approximation based on the fully observable MDP uses Q-functions (Littman, Cassandra, & Kaelbling, 1995):

$$\widehat{V}(b) = \max_{a \in A} \sum_{s \in S} b(s) Q_{MDP}^*(s,a),$$

where

$$Q_{MDP}^*(s,a) = \rho(s,a) + \gamma \sum_{s' \in S} P(s'|s,a) V_{MDP}^*(s')$$

is the optimal action-value function (Q-function) for the fully observable MDP. The QMDP approximation $\widehat{V}$ is piecewise linear and convex with $|A|$ linear functions, each corresponding to one action (Figure 12b). The QMDP update rule (for the belief state MDP) for $\widehat{V}_i$ with linear functions $\alpha_i^k \in \Gamma_i$ is:

$$
\begin{aligned}
\widehat{V}_{i+1}(b) &= \max_{a \in A} \sum_{s \in S} b(s) \left[ \rho(s, a) + \gamma \sum_{s' \in S} P(s'|s, a) \max_{\alpha_i \in \Gamma_i} \alpha_i(s') \right] \\
&= (H_{QMDP} \widehat{V}_i)(b).
\end{aligned}
$$

$H_{QMDP}$ generates a value function with $|A|$ linear functions. The time complexity of the update is the same as for the MDP-approximation case – $O(|A||S|^2 + |\Gamma_i||S|)$, which reduces to $O(|A||S|^2)$ time when only QMDP updates are used. $H_{QMDP}$ is a contraction mapping and its fixed-point solution can be found by solving the corresponding fully observable MDP.

The QMDP update upper-bounds the exact update. The bound is tighter than the MDP update; that is, $H\widehat{V}_i \leq H_{QMDP}\widehat{V}_i \leq H_{MDP}\widehat{V}_i$, as we prove later in Theorem 9. The same inequalities hold for both fixed-point solutions (through Theorem 6).

To illustrate the difference in the quality of bounds for the MDP approximation and the QMDP method, we use our Maze20 navigation problem. To measure the quality of a bound we use the mean of value-function values. Since all belief states are equally important we assume that they are uniformly distributed. We approximate this measure using the average of values for a fixed set of $N = 2000$ belief points. The points in the set were selected uniformly at random in the beginning. Once the set was chosen, it was fixed and remained the same for all tests (here and later). Figure 13 shows the results of the experiment; we include also results for the fast informed bound method that is presented in the next section.[15] Figure 13 also shows the running times of the methods. The methods were implemented in Common Lisp and run on Sun Ultra 1 workstation.

### 4.1.3 CONTROL

The MDP and the QMDP value-function approximations can be used to construct controllers based on one-step lookahead. In addition, the QMDP approximation is also suitable for the direct control strategy, which selects an action corresponding to the best (highest value) Q-function. Thus, the method is a special case of the Q-function approach discussed in Section 3.1.3.[16] The advantage of the direct QMDP method is that it is faster than both lookahead designs. On other the hand, lookahead tends to improve the control performance. This is shown in Figure 14, which compares the control performance of different controllers on the Maze20 problem.

The quality of a policy $\widehat{\pi}$, with no preference towards a particular initial belief state, can be measured by the mean of value-function values for $\widehat{\pi}$ and uniformly distributed initial belief states. We approximate this measure using the average of discounted rewards for 2000 control trajectories obtained for the fixed set of $N = 2000$ initial belief states (selected uniformly at random at the beginning). The trajectories were obtained through simulation and were 60 steps long.[17]

---

15. The confidence interval limits for probability level 0.95 range in $\pm(0.45, 0.62)$ from their respective average scores and this holds for all bound experiments in the paper. As these are relatively small we do not include them in our graphs.

16. As pointed out by Littman et al. (1995), in some instances, the direct QMDP controller never selects investigative actions, that is, actions that try to gain more information about the underlying process state. Note, however, that this observation is not true in general and the QMDP-based controller with direct action selection may select investigative actions, even though in the fully observable version of the problem investigative actions are never chosen.

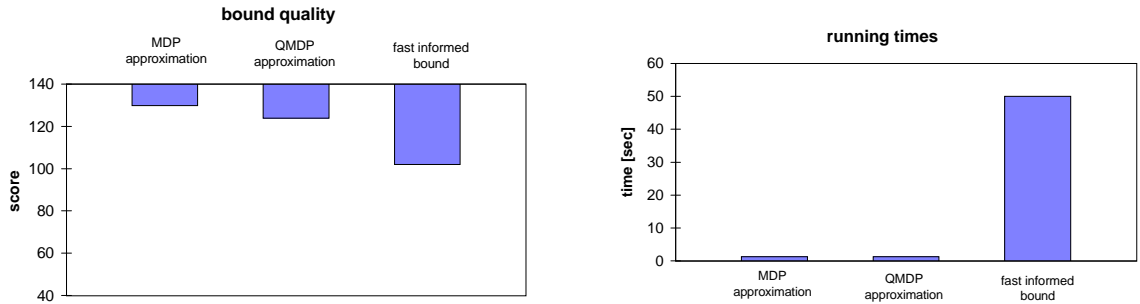

Figure 13: Comparison of the MDP, QMDP and fast informed bound approximations: bound quality (left); running times (right). The bound-quality score is the average value of the approximation for the set of 2000 belief points (chosen uniformly at random). As the methods upper-bound the optimal value function, we flip the bound-quality graph so that longer bars indicate better approximations.

To validate the comparison along the averaged performance scores, we must show that these scores are not the result of randomness and that methods are indeed statistically significantly different. To do this we rely on pairwise significance tests.[18] To summarize the obtained results, the score differences of 1.54, 2.09 and 2.86 between any two methods (here and also later in the paper) are sufficient to reject the method with a lower score being the better performer at significance levels 0.05, 0.01 and 0.001 respectively.[19] Error-bars in Figure 14 reflect the critical score difference for the significance level 0.05.

Figure 14 also shows the average reaction times for different controllers during these experiments. The results show the clear dominance of the direct QMDP controller, which need not do a lookahead in order to extract an action, compared to the other two MDP-based controllers.

## 4.2 Fast Informed Bound Method

Both the MDP and the QMDP approaches ignore partial observability and use the fully observable MDP as a surrogate. To improve these approximations and account (at least to some degree) for partial observability we propose a new method – *the fast informed bound method*. Let $\widehat{V}_i$ be a piecewise linear and convex value function represented by a set of linear functions $\Gamma_i$. The new update is defined as

$$
\begin{aligned}
\widehat{V}_{i+1}(b) &= \max_{a \in A} \left\{ \sum_{s \in S} \rho(s,a)b(s) + \gamma \sum_{o \in \Theta} \sum_{s \in S} \max_{\alpha_i \in \Gamma_i} \sum_{s' \in S} P(s',o|s,a)b(s)\alpha_i(s') \right\} \\
&= \max_{a \in A} \left\{ \sum_{s \in S} b(s) \left[ \rho(s,a) + \gamma \sum_{o \in \Theta} \max_{\alpha_i \in \Gamma_i} \sum_{s' \in S} P(s',o|s,a)\alpha_i(s') \right] \right\} \\
&= (H_{FIB}\widehat{V}_i)(b).
\end{aligned}
$$

---

17. The length of the trajectories (60 steps) for the Maze20 problem was chosen to ensure that our estimates of (discounted) cumulative rewards are not far from the actual rewards for an infinite number of steps.

18. An alternative way to compare two methods is to compute confidence limits for their scores and inspect their overlaps. However, in this case, the ability to distinguish two methods can be reduced due to fluctuations of scores for different initializations. For Maze20, confidence interval limits for probability level 0.95 range in $\pm(1.8, 2.3)$ from their respective average scores. This covers all control experiments here and later. Pairwise tests eliminate the dependency by examining the differences of individual values and thus improve the discriminative power.

19. The critical score differences listed cover the worst case combination. Thus, there may be some pairs for which the smaller difference would suffice.

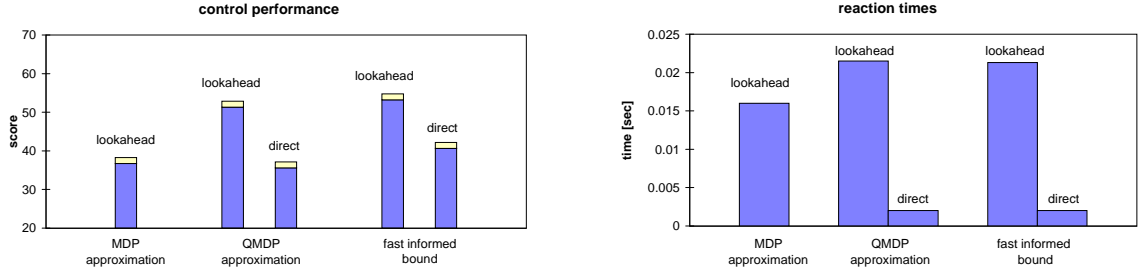

The fast informed bound update can be obtained from the exact update by the following derivation:

$$
\begin{aligned}
(H\widehat{V}_i)(b) &= \max_{a \in A} \left\{ \sum_{s \in S} \rho(s,a)b(s) + \gamma \sum_{o \in \Theta} \max_{\alpha_i \in \Gamma_i} \sum_{s' \in S} \sum_{s \in S} P(s',o|s,a)b(s)\alpha_i(s') \right\} \\
&\leq \max_{a \in A} \left\{ \sum_{s \in S} \rho(s,a)b(s) + \gamma \sum_{o \in \Theta} \sum_{s \in S} \max_{\alpha_i \in \Gamma_i} \sum_{s' \in S} P(s',o|s,a)b(s)\alpha_i(s') \right\} \\
&= \max_{a \in A} \sum_{s \in S} b(s) \left[ \rho(s,a) + \gamma \sum_{o \in \Theta} \max_{\alpha_i \in \Gamma_i} \sum_{s' \in S} P(s',o|s,a)\alpha_i(s') \right] \\
&= \max_{a \in A} \sum_{s \in S} b(s)\alpha_{i+1}^a(s) \\
&= (H_{FIB}\widehat{V}_i)(b).
\end{aligned}
$$

The value function $\widehat{V}_{i+1} = H_{FIB}\widehat{V}_i$ one obtains after an update is piecewise linear and convex and consists of at most $|A|$ different linear functions, each corresponding to one

action

$$\alpha_{i+1}^a(s) = \rho(s, a) + \gamma \sum_{o \in \Theta} \max_{\alpha_i \in \Gamma_i} \sum_{s' \in S} P(s', o|s, a)\alpha_i(s').$$

The $H_{FIB}$ update is efficient and can be computed in $O(|A||S|^2|\Theta||\Gamma_i|)$ time. As the method always outputs $|A|$ linear functions, the computation can be done in $O(|A|^2|S|^2|\Theta|)$ time, when many $H_{FIB}$ updates are strung together. This is a significant complexity reduction compared to the exact approach: the latter can lead to a function consisting of $|A||\Gamma_i|^{|\Theta|}$ linear functions, which is exponential in the number of observations and in the worst case takes $O(|A||S|^2|\Gamma_i|^{|\Theta|})$ time.

As $H_{FIB}$ updates are of polynomial complexity one can find the approximation for the finite-horizon case efficiently. The open issue remains the problem of finding the solution for the infinite-horizon discounted case and its complexity. To address it we establish the following theorem.

**Theorem 8** *A solution for the fast informed bound approximation can be found by solving an MDP with $|S||A||\Theta|$ states, $|A|$ actions and the same discount factor $\gamma$.*

The full proof of the theorem is deferred to the Appendix. The key part of the proof is the construction of an equivalent MDP with $|S||A||\Theta|$ states representing $H_{FIB}$ updates. Since a finite-state MDP can be solved through linear program conversion, the fixed-point solution for the fast informed bound update is computable efficiently.

### 4.2.1 Fast Informed Bound versus Fully-Observable MDP Approximations

The fast informed update upper-bounds the exact update and is tighter than both the MDP and the QMDP approximation updates.

**Theorem 9** *Let $\widehat{V}_i$ corresponds to a piecewise linear convex value function defined by $\Gamma_i$ linear functions. Then $H\widehat{V}_i \leq H_{FIB}\widehat{V}_i \leq H_{QMDP}\widehat{V}_i \leq H_{MDP}\widehat{V}_i$.*

The key trick in deriving the above result is to swap max and sum operators (the proof is in the Appendix) and thus obtain both to the upper-bound inequalities and the subsequent reduction in the complexity of update rules compared to the exact update. This is also shown in Figure 15. The UMDP approximation, also included in Figure 15, is discussed later in Section 4.3. Thus, the difference among the methods boils down to simple mathematical manipulations. Note that the same inequality relations as derived for updates hold also for their fixed-point solutions (through Theorem 6).

Figure 13a illustrates the improvement of the bound over MDP-based approximations on the Maze20 problem. Note, however, that this improvement is paid for by the increased running-time complexity (Figure 13b).

### 4.2.2 Control

The fast informed bound always outputs a piecewise linear and convex function, with one linear function per action. This allows us to build a POMDP controller that selects an action associated with the best (highest value) linear function directly. Figure 14 compares the control performance of the direct and the lookahead controllers to the MDP and the QMDP controllers. We see that the fast informed bound leads not only to tighter bounds but also to improved control on average. However, we stress that currently there is no theoretical underpinning for this observation and thus it may not be true for all belief states and any problem.

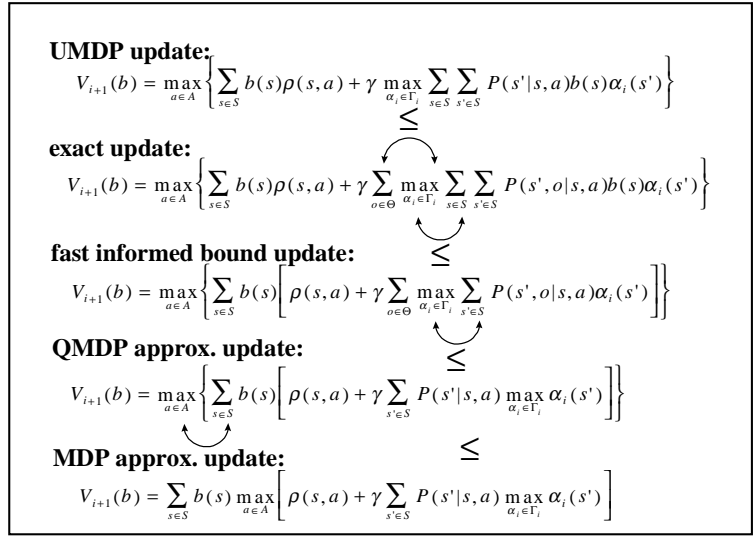

Figure 15: Relations between the exact update and the UMDP, the fast informed bound, the QMDP and the MDP updates.

### 4.2.3 EXTENSIONS OF THE FAST INFORMED BOUND METHOD

The main idea of the fast informed bound method is to select the best linear function for every observation and every current state separately. This differs from the exact update where we seek a linear function that gives the best result for every observation and the combination of all states. However, we observe that there is a great deal of middle ground between these two extremes. Indeed, one can design an update rule that chooses optimal (maximal) linear functions for disjoint sets of states separately. To illustrate this idea, assume a partitioning $\mathcal{S} = \{S_1, S_2, \cdots, S_m\}$ of the state space $S$. The new update for $\mathcal{S}$ is:

$$
\widehat{V}_{i+1}(b) = \max_{a \in A} \left\{ \sum_{s \in S} \rho(s,a)b(s) + \gamma \sum_{o \in \Theta} \left[ \max_{\alpha_i \in \Gamma_i} \sum_{s \in S_1} \sum_{s' \in S} P(s',o|s,a)b(s)\alpha_i(s') + \right.\right.
$$
$$
\max_{\alpha_i \in \Gamma_i} \sum_{s \in S_2} \sum_{s' \in S} P(s',o|s,a)b(s)\alpha_i(s') + \cdots +
$$
$$
\left.\left. \max_{\alpha_i \in \Gamma_i} \sum_{s \in S_m} \sum_{s' \in S} P(s',o|s,a)b(s)\alpha_i(s') \right] \right\}
$$

It is easy to see that the update upper-bounds the exact update. Exploration of this approach and various partitioning heuristics remains an interesting open research issue.

## 4.3 Approximation with Unobservable MDP

The MDP-approximation assumes full observability of POMDP states to obtain simpler and more efficient updates. The other extreme is to discard all observations available to the decision maker. An MDP with no observations is called *unobservable MDP (UMDP)* and one may choose its value-function solution as an alternative approximation.

To find the solution for the unobservable MDP, we derive the corresponding update rule, $H_{UMDP}$, similarly to the update for the partially observable case. $H_{UMDP}$ preserves piecewise linearity and convexity of the value function and is a contraction. The update equals:

$$\widehat{V}_{i+1}(b) = \max_{a \in A} \left\{ \sum_{s \in S} \rho(s,a)b(s) + \gamma \max_{\alpha_i \in \Gamma_i} \sum_{s \in S} \sum_{s' \in S} P(s'|s,a)b(s)\alpha_i(s') \right\}$$
$$= (H_{UMDP}\widehat{V}_i)(b),$$

where $\Gamma_i$ is a set of linear functions describing $\widehat{V}_i$. $\widehat{V}_{i+1}$ remains piecewise linear and convex and it consists of at most $|\Gamma_i||A|$ linear functions. This is in contrast to the exact update, where the number of possible vectors in the next step can grow exponentially in the number of observations and leads to $|A||\Gamma_i|^{|\Theta|}$ possible vectors. The time complexity of the update is $O(|A||S|^2|\Gamma_i|)$. Thus, starting from $\widehat{V}_0$ with one linear function, the running-time complexity for $k$ updates is bounded by $O(|A|^k|S|^2)$. The problem of finding the optimal solution for the unobservable MDP remains intractable: the finite-horizon case is NP-hard(Burago et al., 1996), and the discounted infinite-horizon case is undecidable (Madani et al., 1999). Thus, it is usually not very useful approximation.

The update $H_{UMDP}$ lower-bounds the exact update, an intuitive result reflecting the fact that one cannot do better with less information. To provide some insight into how the two updates are related, we do the following derivation, which also proves the bound property in an elegant way:

$$(H\widehat{V}_i)(b) = \max_{a \in A} \left\{ \sum_{s \in S} \rho(s,a)b(s) + \gamma \sum_{o \in \Theta} \max_{\alpha_i \in \Gamma_i} \sum_{s' \in S} \sum_{s \in S} P(s',o|s,a)b(s)\alpha_i(s') \right\}$$
$$\geq \max_{a \in A} \left\{ \sum_{s \in S} \rho(s,a)b(s) + \gamma \max_{\alpha_i \in \Gamma_i} \sum_{o \in \Theta} \sum_{s \in S} \sum_{s' \in S} P(s',o|s,a)b(s)\alpha_i(s') \right\}$$
$$= \max_{a \in A} \left\{ \sum_{s \in S} \rho(s,a)b(s) + \gamma \max_{\alpha_i \in \Gamma_i} \sum_{s \in S} \sum_{s' \in S} P(s'|s,a)b(s)\alpha_i(s') \right\}$$
$$= (H_{UMDP}\widehat{V}_i)(b).$$

We see that the difference between the exact and UMDP updates is that the max and the sum over next-step observations are exchanged. This causes the choice of $\alpha$ vectors in $H_{UMDP}$ to become independent of the observations. Once the sum and max operations are exchanged, the observations can be marginalized out. Recall that the idea of swaps leads to a number of approximation updates; see Figure 15 for their summary.

## 4.4 Fixed-Strategy Approximations

A finite-state machine (FSM) model is used primarily to define a control strategy. Such a strategy does not require belief state updates since it directly maps sequences of observations to sequences of actions. The value function of an FSM strategy is piecewise linear and convex and can be found efficiently in the number of memory states (Section 2.6.1). While in the policy iteration and policy approximation contexts the value function for a specific strategy is used to quantify the goodness of the policy in the first place, the value function alone can be also used as a substitute for the optimal value function. In this case, the value function (defined over the belief space) equals

$$V^C(b) = \max_{x \in M} V^C(x, b),$$

where $V^C(x, b) = \sum_{s \in S} V^C(x, s) b(s)$ is obtained by solving a set of $|S||M|$ linear equations (Section 2.6.2). As remarked earlier, the value for the fixed strategy lower-bounds the optimal value function, that is $V^C \leq V^*$.

To simplify the comparison of the fixed-strategy approximation to other approximations, we can rewrite its solution also in terms of fixed-strategy updates

$$
\begin{aligned}
\widehat{V}_{i+1}(b) &= \max_{x \in M} \left\{ \sum_{s \in S} \rho(s, \eta(x)) b(s) + \gamma \sum_{o \in \Theta} \sum_{s \in S} \sum_{s' \in S} P(o, s'|s, \eta(x)) b(s) \alpha_i(\phi(x, o), s') \right\}, \\
&= \max_{x \in M} \left\{ \sum_{s \in S} b(s) \left[ \rho(s, \eta(x)) + \gamma \sum_{o \in \Theta} \sum_{s' \in S} P(o, s'|s, \eta(x)) \alpha_i(\phi(x, o), s') \right] \right\} \\
&= (H_{FSM} \widehat{V}_i)(b).
\end{aligned}
$$

The value function $\widehat{V}_i$ is piecewise linear and convex and consists of $|M|$ linear functions $\alpha_i(x, .)$. For the infinite-horizon discounted case $\alpha_i(x, s)$ represents the $i$th approximation of $V^C(x, s)$. Note that the update can be applied to the finite-horizon case in a straightforward way.

### 4.4.1 QUALITY OF CONTROL

Assume we have an FSM strategy and would like to use it as a substitute for the optimal control policy. There are three different ways in which we can use it to extract the control. The first is to simply execute the strategy represented by the FSM. There is no need to update belief states in this case. The second possibility is to choose linear functions corresponding to different memory states and their associated actions repeatedly in every step. We refer to such a controller as a *direct* (DR) controller. This approach requires updating of belief states in every step. On the other hand its control performance is no worse than that of the FSM control. The final strategy discards all the information about actions and extracts the policy by using the value function $\widehat{V}(b)$ and one-step lookahead. This method (LA) requires both belief state updates and lookaheads and leads to the worst reactive time. Like DR, however, this strategy is guaranteed to be no worse than the FSM controller. The following theorem relates the performances of the three controllers.

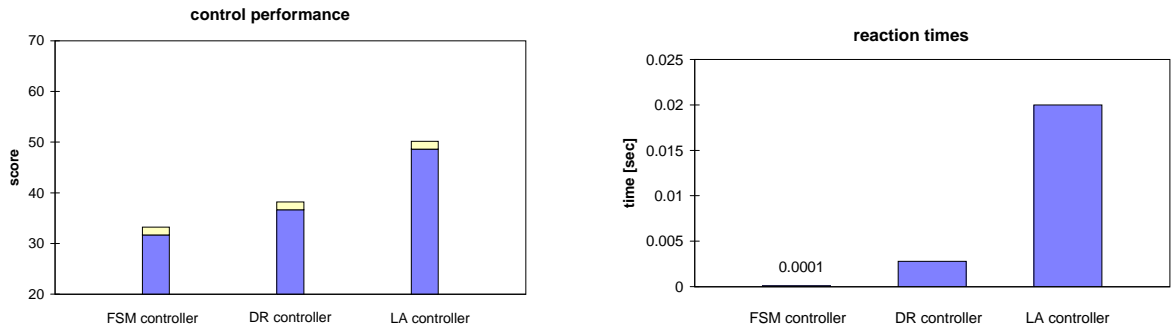

Figure 16: Comparison of three different controllers (FSM, DR and LA) for the Maze20 problem and a collection of one-action policies: control quality (left) and response time (right). Error-bars in the control performance graph indicate the critical score difference at which any two methods become statistically different at significance level 0.05.

**Theorem 10** *Let $C_{FSM}$ be an FSM controller. Let $C_{DR}$ and $C_{LA}$ be the direct and the one-step-lookahead controllers constructed based on $C_{FSM}$. Then $V^{C_{FSM}}(b) \leq V^{C_{DR}}(b)$ and $V^{C_{FSM}}(b) \leq V^{C_{LA}}(b)$ hold for all belief states $b \in \mathcal{I}$.*

Though we can prove that both the direct controller and the lookahead controller are always better than the underlying FSM controller (see Appendix for the full proof of the theorem), we cannot show the similar property between the first two controllers for all initial belief states. However, the lookahead approach typically tends to dominate, reflecting the usual trade-off between control quality and response time. We illustrate this trade-off on our running Maze20 example and a collection of $|A|$ one-action policies, each generating a sequence of the same action. Control quality and response time results are shown in Figure 16. We see that the controller based on the FSM is the fastest of the three, but it is also the worst in terms of control quality. On the other hand, the direct controller is slower (it needs to update belief states in every step) but delivers better control. Finally, the lookahead controller is the slowest and has the best control performance.

### 4.4.2 Selecting the FSM Model

The quality of a fixed-strategy approximation depends strongly on the FSM model used. The model can be provided a priori or constructed automatically. Techniques for automatic construction of FSM policies correspond to a search problem in which either the complete or a restricted space of policies is examined to find the optimal or the near-optimal policy for such a space. The search process is equivalent to policy approximations or policy-iteration techniques discussed earlier in Sections 2.6 and 3.2.

### 4.5 Grid-Based Approximations with Value Interpolation-Extrapolation

A value function over a continuous belief space can be approximated by a finite set of grid points $G$ and an *interpolation-extrapolation rule* that estimates the value of an arbitrary point of the belief space by relying only on the points of the grid and their associated values.

**Definition 8** (*Interpolation-extrapolation rule*) *Let* $f : \mathcal{I} \to \mathbb{R}$ *be a real-valued function defined over the information space* $\mathcal{I}$, $G = \{b_1^G, b_2^G, \cdots b_k^G\}$ *be a set of grid points and* $\Psi^G = \{(b_1^G, f(b_1^G)), (b_2^G, f(b_2^G)), \cdots, (b_k^G, f(b_k^G))\}$ *be a set of point-value pairs. A function* $R_G : \mathcal{I} \times (\mathcal{I} \times \mathbb{R})^{|G|} \to \mathbb{R}$ *that estimates* $f$ *at any point of the information space* $\mathcal{I}$ *using only values associated with grid points is called an interpolation-extrapolation rule.*

The main advantage of an interpolation-extrapolation model in estimating the true value function is that it requires us to compute value updates only for a finite set of grid points $G$. Let $\widehat{V}_i$ be the approximation of the $i$th value function. Then the approximation for the $(i+1)$th value function $\widehat{V}_{i+1}$ can be obtained as

$$\widehat{V}_{i+1}(b) = R_G(b, \Psi_{i+1}^G),$$

where values associated with every grid point $b_j^G \in G$ (and included in $\Psi_{i+1}^G$) are:

$$\varphi_{i+1}(b_j^G) = (H\widehat{V}_i)(b_j^G) = \max_{a \in A} \left\{ \rho(b,a) + \gamma \sum_{o \in \Theta} P(o|b,a)\widehat{V}_i(\tau(b_j^G, o, a)) \right\}. \tag{9}$$

The grid-based update can also be described in terms of a value-function mapping $H_G$: $\widehat{V}_{i+1} = H_G\widehat{V}_i$. The complexity of such an update is $O(|G||A||S|^2|\Theta|C_{\text{Eval}}(R_G, |G|))$ where $C_{\text{Eval}}(R_G, |G|)$ is the computational cost of evaluating the interpolation-extrapolation rule $R_G$ for $|G|$ grid points. We show later (Section 4.5.3), that in some instances, the need to evaluate the interpolation-extrapolation rule in every step can be eliminated.

#### 4.5.1 A FAMILY OF CONVEX RULES

The number of all possible interpolation-extrapolation rules is enormous. We focus on a set of *convex rules* that is a relatively small but very important subset of interpolation-extrapolation rules.[20]

**Definition 9** (*Convex rule*) *Let* $f$ *be some function defined over the space* $\mathcal{I}$, $G = \{b_1^G, b_2^G, \cdots b_k^G\}$ *be a set of grid points, and* $\Psi^G = \{(b_1^G, f(b_1^G)), (b_2^G, f(b_2^G)), \cdots, (b_k^G, f(b_k^G))\}$ *be a set of point-value pairs. The rule* $R_G$ *for estimating* $f$ *using* $\Psi^G$ *is called convex when for every* $b \in \mathcal{I}$, *the value* $\widehat{f}(b)$ *is:*

$$\widehat{f}(b) = R_G(b, \Psi^G) = \sum_{j=1}^{|G|} \lambda_j^b f(b_j),$$

*such that* $0 \le \lambda_j^b \le 1$ *for every* $j = 1, \cdots, |G|$, *and* $\sum_{j=1}^{|G|} \lambda_j^b = 1$.

---

20. We note that convex rules used in our work are a special case of averagers introduced by Gordon (1995). The difference is minor; the definition of an averager includes a constant (independent of grid points and their values) that is added to the convex combination.

The key property of convex rules is that their corresponding grid-based update $H_G$ is a contraction in the max norm (Gordon, 1995). Thus, the approximate value iteration based on $H_G$ converges to the unique fixed-point solution. In addition, $H_G$ based on convex rules is isotone.

### 4.5.2 EXAMPLES OF CONVEX RULES

The family of convex rules includes approaches that are very commonly used in practice, like *nearest neighbor, kernel regression, linear point interpolations* and many others.

Take, for example, the *nearest-neighbor* approach. The function for a belief point $b$ is estimated using the value at the grid point closest to it in terms of some distance metric $M$ defined over the belief space. Then, for any point $b$, there is exactly one nonzero parameter $\lambda_j^b = 1$ such that $\| b - b_j^G \|_M \leq \| b - b_i^G \|_M$ holds for all $i = 1, 2, \cdots, k$. All other $\lambda$s are zero. Assuming the Euclidean distance metric, the nearest-neighbor approach leads to a piecewise constant approximation, in which regions with equal values correspond to regions with a common nearest grid point.

The nearest neighbor estimates the function value by taking into an account only one grid point and its value. *Kernel regression* expands upon this by using more grid points. It adds up and weights their contributions (values) according to their distance from the target point. For example, assuming Gaussian kernels, the weight for a grid point $b_j^G$ is

$$\lambda_j^b = \beta \exp^{-\|b - b_j^G\|_M^2 / 2\sigma^2},$$

where $\beta$ is a normalizing constant ensuring that $\sum_{j=1}^{|G|} \lambda_j^b = 1$ and $\sigma$ is a parameter that flattens or narrows weight functions. For the Euclidean metric, the above kernel-regression rule leads to a smooth approximation of the function.

*Linear point interpolations* are a subclass of convex rules that in addition to constraints in Definition 9 satisfy

$$b = \sum_{j=1}^{|G|} \lambda_j^b b_j^G.$$

That is, a belief point $b$ is a convex combination of grid points and the $\lambda$s are the corresponding coefficients. Because the optimal value function for the POMDP is convex, the new constraint is sufficient to prove the upper-bound property of the approximation. In general, there can be many different linear point-interpolations for a given grid. A challenging problem here is to find the rule with the best approximation. We discuss these issues in Section 4.5.7.

### 4.5.3 CONVERSION TO A GRID-BASED MDP

Assume that we would like to find the approximation of the value function using our grid-based convex rule and grid-based update (Equation 9). We can view this process also as a process of finding a sequence of values $\varphi_1(b_j^G), \varphi_2(b_j^G), \cdots, \varphi_i(b_j^G), \cdots$ for all grid-points $b_j^G \in G$. We show that in some instances the sequence of values can be computed without applying an interpolation-extrapolation rule in every step. In such cases, the problem can be converted into a fully observable MDP with states corresponding to grid-points $G$.[21] We call this MDP a *grid-based MDP*.

**Theorem 11** *Let $G$ be a finite set of grid points and $R_G$ be a convex rule such that parameters $\lambda_j^b$ are fixed. Then the values of $\varphi(b_j^G)$ for all $b_j^G \in G$ can be found by solving a fully observable MDP with $|G|$ states and the same discount factor $\gamma$.*

**Proof** For any grid point $b_j^G$ we can write:

$$
\begin{aligned}
\varphi_{i+1}(b_j^G) &= \max_{a \in A} \left\{ \rho(b_j^G, a) + \gamma \sum_{o \in \Theta} P(o|b_j^G, a) \widehat{V}_i^G(\tau(b_j^G, a, o)) \right\} \\
&= \max_{a \in A} \left\{ \rho(b_j^G, a) + \gamma \sum_{o \in \Theta} P(o|b_j^G, a) \left[ \sum_{k=1}^{|G|} \lambda_{j,k}^{o,a} \varphi_i(b_k^G) \right] \right\} \\
&= \max_{a \in A} \left\{ \left[ \rho(b_j^G, a) \right] + \gamma \sum_{k=1}^{|G|} \varphi_i^G(b_k^G) \left[ \sum_{o \in \Theta} P(o|b_j^G, a) \lambda_{j,k}^{o,a} \right] \right\}
\end{aligned}
$$

Now denoting $\left[ \sum_{o \in \Theta} P(o|b_j, a)^G \lambda_{j,k}^{o,a} \right]$ as $P(b_k^G | b_j^G, a)$, we can construct a fully observable MDP problem with states corresponding to grid points $G$ and the same discount factor $\gamma$. The update step equals:

$$
\varphi_{i+1}(b_j^G) = \max_{a \in A} \left\{ \rho(b_j^G, a) + \gamma \sum_{k=1}^{|G|} P(b_k^G | b_j^G, a) \varphi_i^G(b_k^G) \right\}.
$$

The prerequisite $0 \le \lambda_j^b \le 1$ for every $j = 1, \cdots, |G|$ and $\sum_{j=1}^{|G|} \lambda_j^b = 1$ guarantees that $P(b_k^G | b_j^G, a)$ can be interpreted as true probabilities. Thus, one can compute values $\varphi(b_j^G)$ by solving the equivalent fully-observable MDP. $\quad\square$

### 4.5.4 SOLVING GRID-BASED APPROXIMATIONS

The idea of converting a grid-based approximation into a grid-based MDP is a basis of our simple but very powerful approximation algorithm. Briefly, the key here is to find the parameters (transition probabilities and rewards) of a new MDP model and then solve it. This process is relatively easy if the parameters $\lambda$ used to interpolate-extrapolate the value of a non-grid point are fixed (the assumption of Theorem 11). In such a case, we can determine parameters of the new MDP efficiently in one step, for any grid set $G$. The nearest neighbor or the kernel regression are examples of rules with this property. Note that this leads to polynomial-time algorithms for finding values for all grid points (recall that an MDP can be solved efficiently for both finite and discounted, infinite-horizon criteria).

The problem in solving grid-based approximation arises only when the parameters $\lambda$ used in the interpolation-extrapolation are not fixed and are subject to the optimization itself. This happens, for example, when there are multiple ways of interpolating a value at some point of the belief space and we would like to find the best interpolation (leading to the best values) for all grid points in $G$. In such a case, the corresponding "optimal" grid-based MDP cannot be found in a single step and iterative approximation, solving a sequence of grid-based MDPs, is usually needed. The worst-case complexity of this problem remains an open question.

---

21. We note that a similar result has been also proved independently by Gordon (1995).

### 4.5.5 CONSTRUCTING GRIDS

An issue we have not touched on so far is the selection of grids. There are multiple ways to select grids. We divide them into two classes – regular and non-regular grids.

Regular grids (Lovejoy, 1991a) partition the belief space evenly into equal-size regions.[22] The main advantage of regular grids is the simplicity with which we can locate grid points in the neighborhood of any belief point. The disadvantage of regular grids is that they are restricted to a specific number of points, and any increase in grid resolution is paid for in an exponential increase in the grid size. For example, a sequence of regular grids for a 20-dimensional belief space (corresponds to a POMDP with 20 states) consists of 20, 210, 1540, 8855, 42504, $\cdots$ grid points.[23] This prevents one from using the method with higher grid resolutions for problems with larger state spaces.

Non-regular grids are unrestricted and thus provide for more flexibility when grid resolution must be increased adaptively. On the other hand, due to irregularities, methods for locating grid points adjacent to an arbitrary belief point are usually more complex when compared to regular grids.

### 4.5.6 LINEAR POINT INTERPOLATION

The fact that the optimal value function $V^*$ is convex for a belief-state MDPs can be used to show that the approximation based on linear point interpolation always upper-bounds the exact solution (Lovejoy, 1991a, 1993). Neither kernel regression nor nearest neighbor can guarantee us any bound.

**Theorem 12** (*Upper bound property of a grid-based point interpolation update*). *Let $\widehat{V}_i$ be a convex value function. Then $H\widehat{V}_i \leq H_G\widehat{V}_i$.*

The upper-bound property of $H_G$ update for convex value functions follows directly from Jensen's inequality. The convergence to an upper-bound follows from Theorem 6.

Note that the point-interpolation update imposes an additional constraint on the choice of grid points. In particular, it is easy to see that any valid grid must also include extreme points of the belief simplex (extreme points correspond to $(1, 0, 0, \cdots), (0, 1, 0, \cdots)$,

---

22. Regular grids used by Lovejoy (1991a) are based on Freudenthal triangulation (Eaves, 1984). Essentially, this is the same idea as used to partition evenly the n-dimensional subspace of $I\!\!R^n$. In fact, an affine transform allows us to map isomorphically grid points in the belief space to grid points in the $n$-dimensional space (Lovejoy, 1991a).

23. The number of points in the regular grid sequence is given by (Lovejoy, 1991a):

$$|G| = \frac{(M + |S| - 1)!}{M!(|S| - 1)!},$$

where $M = 1, 2, \cdots$ is a grid refinement parameter.

etc.). Without extreme points one would be unable to cover the whole belief space via interpolation. Nearest neighbor and kernel regression impose no restrictions on the grid.

### 4.5.7 FINDING THE BEST INTERPOLATION

In a general, there are multiple ways to interpolate a point of a belief space. Our objective is to find the best interpolation, that is, the one that leads to the tightest upper bound of the optimal value function.

Let $b$ be a belief point and $\{(b_j, f(b_j))|b_j \in G\}$ a set of grid-value pairs. Then the best interpolation for point $b$ is:

$$\widehat{f}(b) = \min_{\overline{\lambda}} \sum_{j=1}^{|G|} \lambda_j f(b_j)$$

subject to $0 \leq \lambda_j \leq 1$ for all $j = 1, \cdots, |G|$, $\sum_{j=1}^{|G|} \lambda_j = 1$, and $b = \sum_{j=1}^{|G|} \lambda_j b_j^G$.

This is a linear optimization problem. Although it can be solved in polynomial time (using linear programming techniques), the computational cost of doing this is still relatively large, especially considering the fact that the optimization must be repeated many times. To alleviate this problem we seek more efficient ways of finding the interpolation, sacrificing the optimality.

One way to find a (suboptimal) interpolation quickly is to apply regular grids proposed by Lovejoy (1991a). In this case the value at a belief point is approximated using the convex combination of grid points closest to it. The approximation leads to piecewise linear and convex value functions. As all interpolations are fixed here, the problem of finding the approximation can be converted into an equivalent grid-based MDP and solved as a finite-state MDP. However, as pointed in the previous section, the regular grids must use a specific number of grid points and any increase in the resolution of a grid is paid for by an exponential increase in the grid size. This feature makes the method less attractive when we have a problem with a large state space and we need to achieve high grid resolution.[24]

In the present work we focus on non-regular (or arbitrary) grids. We propose an interpolation approach that searches a limited space of interpolations and is guaranteed to run in time linear in the size of the grid. The idea of the approach is to interpolate a point $b$ of a belief space of dimension $|S|$ with a set of grid points that consists of an arbitrary grid point $b' \in G$ and $|S| - 1$ extreme points of the belief simplex. The coefficients of this interpolation can be found efficiently and we search for the best such interpolation. Let $b' \in G$ be a grid point defining one such interpolation. Then the value at point $b$ satisfies

$$\widehat{V}_i(b) = \min_{b' \in G} \widehat{V}_i^{b'}(b),$$

where $\widehat{V}_i^{b'}$ is the value of the interpolation for the grid point $b'$. Figure 17 illustrates the resulting approximation. The function is characterized by its "sawtooth" shape, which is influenced by the choice of the interpolating set.

To find the best value-function solution or its close approximation we can apply a value iteration procedure in which we search for the best interpolation after every update step.

---

24. One solution to this problem may be to use adaptive regular grids in which grid resolution is increased only in some parts of the belief space. We leave this idea for future work.

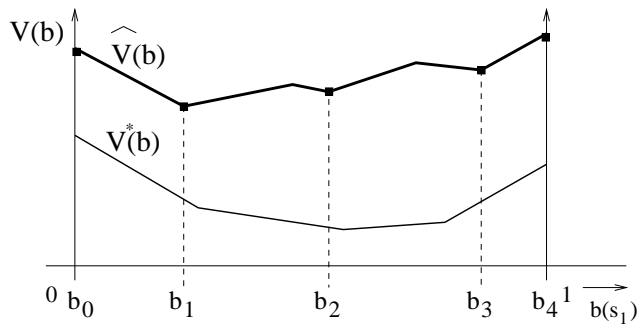

Figure 17: Value-function approximation based on the linear-time interpolation approach (a two-dimensional case). Interpolating sets are restricted to a single internal point of the belief space.

The drawback of this approach is that interpolations may remain unchanged for many update steps, thus slowing down the solution process. An alternative approach is to solve a sequence of grid-based MDPs instead. In particular, at every stage we find the best (minimum value) interpolations for all belief points reachable from grid points in one step, fix coefficients of these interpolations ($\lambda$s), construct a grid-based MDP and solve it (exactly or approximately). This process is repeated until no further improvement (or no improvement larger than some threshold) is seen in values at different grid points.

### 4.5.8 IMPROVING GRIDS ADAPTIVELY

The quality of an approximation (bound) depends strongly on the points used in the grid. Our objective is to provide a good approximation with the smallest possible set of grid points. However, this task is impossible to achieve, since it cannot be known in advance (before solving) what belief points to pick. A way to address this problem is to build grids incrementally, starting from a small set of grid points and adding others adaptively, but only in places with a greater chance of improvement. The key part of this approach is a heuristic for choosing grid points to be added next.

One heuristic method we have developed attempts to maximize improvements in bound values via stochastic simulations. The method builds on the fact that every interpolation grid must also include extreme points (otherwise we cannot cover the entire belief space). As extreme points and their values affect the other grid points, we try to improve their values in the first place. In general, a value at any grid point $b$ improves more the more precise values are used for its successor belief points, that is, belief states that correspond to $\tau(b, a^*, o)$ for a choice of observation $o$. $a^*$ is the current optimal action choice for $b$. Incorporating such points into the grid then makes a larger improvement in the value at the initial grid point $b$ more likely. Assuming that our initial point is an extreme point, we have a heuristic that tends to improve a value for that point. Naturally, one can proceed further with this selection by incorporating the successor points for the first-level successors into the grid as well, and so forth.

**generate new grid points** $(G, \widehat{V}^G)$
    **set** $G_{new} = \{\}$
    **for** all extreme points $b$ do
        **repeat** until $b \notin G \cup G_{new}$
            **set** $a^* = \arg\max_a \left\{ \rho(b,a) + \gamma \sum_{o \in \Theta} P(o|b,a) \widehat{V}^G(\tau(b,a,o)) \right\}$
            **select** observation $o$ according to $P(o|b,a^*)$
            **update** $b = \tau(b,a^*,o)$
        **add** $b$ into $G_{new}$
    **return** $G_{new}$

Figure 18: Procedure for generating additional grid points based on our bound improvement heuristic.

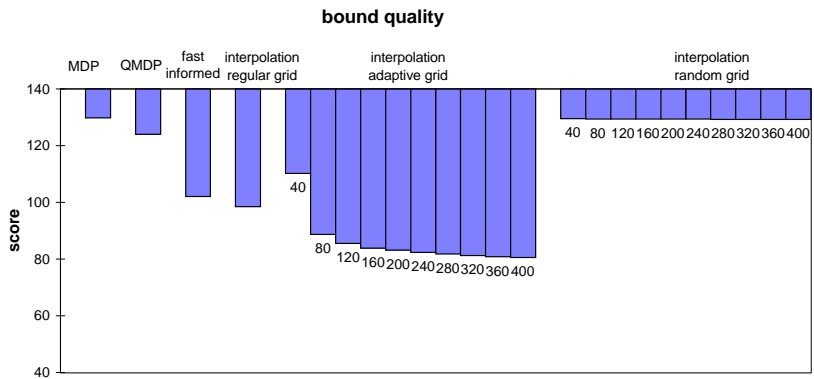

Figure 19: Improvement in the upper bound quality for grid-based point-interpolations based on the adaptive-grid method. The method is compared to randomly refined grid and the regular grid with 210 points. Other upper-bound approximations (the MDP, QMDP and fast informed bound methods) are included for comparison.

To capture this idea, we generate new grid points via simulation, starting from one of the extremes of the belief simplex and continuing until a belief point not currently in the grid is reached. An algorithm that implements the bound improvement heuristic and expands the current grid $G$ with a set of $|S|$ new grid points while relying on the current value-function approximation $\widehat{V}^G$ is shown in Figure 18.

Figure 19 illustrates the performance (bound quality) of our adaptive grid method on the Maze20 problem. Here we use the combination of adaptive grids with our linear-time interpolation approach. The method gradually expands the grid in 40 point increments up to 400 grid points. Figure 19 also shows the performance of the random-grid method in which new points of the grid are selected iniformly at random (results for 40 grid point increments are shown). In addition, the figure gives results for the regular grid interpolation (based on Lovejoy (1991a)) with 210 belief points and other upper-bound methods: the MDP, the QMDP and the fast informed bound approximations.

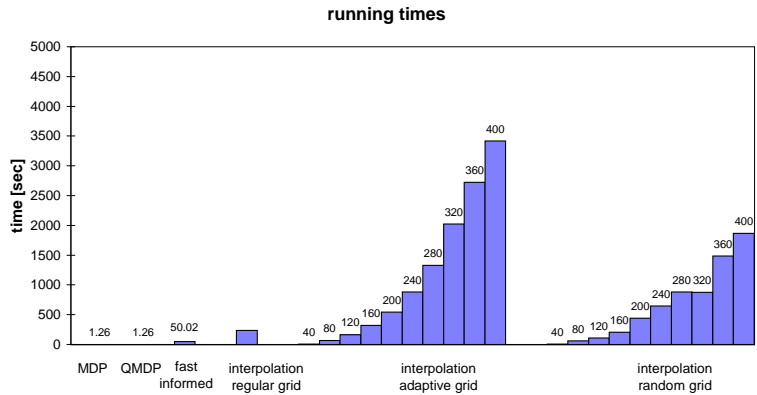

Figure 20: Running times of grid-based point-interpolation methods. Methods tested include the adaptive grid, the random grid, and the regular grid with 210 grid points. Running times for the adaptive-grid are cumulative, reflecting the dependencies of higher grid resolutions on the lower-level resolutions. The running time results for the MDP, QMDP, and fast informed bound approximations are shown for comparison.

We see a dramatic improvement in the quality of the bound for the adaptive method. In contrast to this, the uniformly sampled grid (random-grid approach) hardly changes the bound. There are two reasons for this: (1) uniformly sampled grid points are more likely to be concentrated in the center of the belief simplex; (2) the transition matrix for the Maze20 problem is relatively sparse, the belief points one obtains from the extreme points in one step are on the boundary of the simplex. Since grid points in the center of the simplex are never used to interpolate belief states reachable from extremes in one step they cannot improve values at extremes and the bound does not change.

One drawback of the adaptive method is its running time (for every grid size we need to solve a sequence of grid-based MDPs). Figure 20 compares running times of different methods on the Maze20 problem. As grid-expansion of the adaptive method depends on the value function obtained for previous steps, we plot its cumulative running times. We see a relatively large increase in running time, especially for larger grid sizes, reflecting the trade-off between the bound quality and running time. However, we note that the adaptive-grid method performs quite well in the initial few steps, and with only 80 grid points outperforms the regular grid (with 210 points) in bound quality.

Finally, we note that other heuristic approaches to constructing adaptive grids for point interpolation are possible. For example, a different approach that refines the grid by ex-

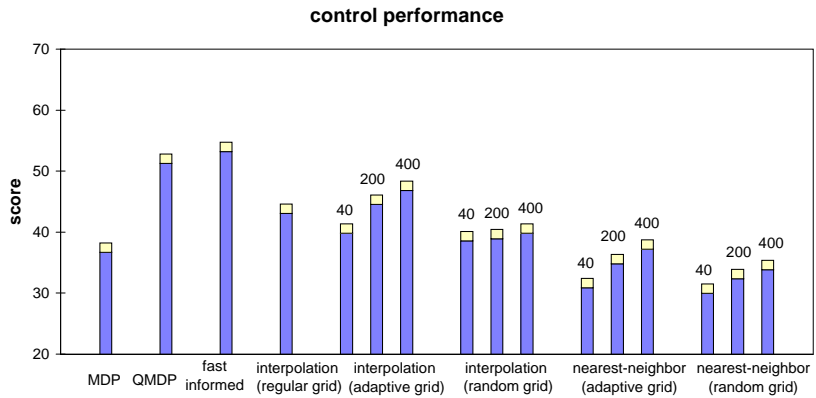

Figure 21: Control performance of lookahead controllers based on grid-based point inter-
polation and nearest neighbor methods and varying grid sizes. The results are
compared to the MDP, the QMDP and the fast informed bound controllers.

amining differences in values at current grid points has recently been proposed by Brafman
(1997).

### 4.5.9 CONTROL

Value functions obtained for different grid-based methods define a variety of controllers. Fig-
ure 21 compares the performances of lookahead controllers based on the point-interpolation
and nearest-neighbor methods. We run two versions of both approaches, one with the adap-
tive grid, the other with the random grid, and we show results obtained for 40, 200 and 400
grid points. In addition, we compare their performances to the interpolation with regular
grids (with 210 grid points), the MDP, the QMDP and the fast informed bound approaches.

Overall, the performance of the interpolation-extrapolation techniques we tested on
the Maze20 problem was a bit disappointing. In particular, better scores were achieved
by the simpler QMDP and fast informed bound methods. We see that, although heuristics
improved the bound quality of approximations, they did not lead to the similar improvement
over the QMDP and the fast informed bound methods in terms of control. This result
shows that a bad bound (in terms of absolute values) does not always imply bad control
performance. The main reason for this is that the control performance is influenced mostly
by relative rather than absolute value-function values (or, in other words, by the shape
of the function). All interpolation-extrapolation techniques we use (except regular grid
interpolation) approximate the value function with functions that are not piecewise linear
and convex; the interpolations are based on the linear-time interpolation technique with a
sawtooth-shaped function, and the nearest-neighbor leads to a piecewise-constant function.
This does not allow them to match the shape of the optimal function correctly. The other
factor that affects the performance is a large sensitivity of methods to the selection of grid
points, as documented, for example, by the comparison of heuristic and random grids.

In the above tests we focused on lookahead controllers only. However, an alternative way to define a controller for grid-based interpolation-extrapolation methods is to use Q-function approximations instead of value functions, and either direct or lookahead designs.[25] Q-function approximations can be found by solving the same grid-based MDP, and by keeping values (functions) for different actions separate at the end.

## 4.6 Approximations of Value Functions Using Curve Fitting (Least-Squares Fit)

An alternative way to approximate a function over a continuous space is to use curve-fitting techniques. This approach relies on a predefined parametric model of the value function and a set of values associated with a finite set of (grid) belief points $G$. The approach is similar to interpolation-extrapolation techniques in that it relies on a set of belief-value pairs. The difference is that the curve fitting, instead of remembering all belief-value pairs, tries to summarize them in terms of a given parametric function model. The strategy seeks the best possible match between model parameters and observed point values. The best match can be defined using various criteria, most often the least-squares fit criterion, where the objective is to minimize

$$Error(f) = \frac{1}{2} \sum_j \left[ y_j - f(b_j) \right]^2.$$

Here $b_j$ and $y_j$ correspond to the belief point and its associated value. The index $j$ ranges over all points of the sample set $G$.

### 4.6.1 COMBINING DYNAMIC PROGRAMMING AND LEAST-SQUARES FIT

The least-squares approximation of a function can be used to construct a dynamic-programming algorithm with an update step: $\widehat{V}_{i+1} = H_{LSF} \widehat{V}_i$. The approach has two steps. First, we obtain new values for a set of sample points $G$:

$$\varphi_{i+1}(b) = (H\widehat{V}_i)(b) = \max_{a \in A} \left\{ \sum_{s \in S} \rho(s,a)b(s) + \gamma \sum_{o \in \Theta} \sum_{s \in S} P(o|s,a)b(s)\widehat{V}_i(\tau(b,a,o)) \right\}.$$

Second, we fit the parameters of the value-function model $\widehat{V}_{i+1}$ using new sample-value pairs and the square-error cost function. The complexity of the update is $O(|G||A||S|^2|\Theta|C_{\text{Eval}}(\widehat{V}_i) + C_{\text{Fit}}(\widehat{V}_{i+1}, |G|))$ time, where $C_{\text{Eval}}(\widehat{V}_i)$ is the computational cost of evaluating $\widehat{V}_i$ and $C_{\text{Fit}}(\widehat{V}_{i+1}, |G|)$ is the cost of fitting parameters of $\widehat{V}_{i+1}$ to $|G|$ belief-value pairs.

The advantage of the approximation based on the least-squares fit is that it requires us to compute updates only for the finite set of belief states. The drawback of the approach is that, when combined with the value-iteration method, it can lead to instability and/or divergence. This has been shown for MDPs by several researchers (Bertsekas, 1994; Boyan & Moore, 1995; Baird, 1995; Tsitsiklis & Roy, 1996).

---

25. This is similar to the QMDP method, which allows both lookahead and greedy designs. In fact, QMDP can be viewed as a special case of the grid-based method with Q-function approximations, where grid points correspond to extremes of the belief simplex.

### 4.6.2 On-line Version of the Least-Squares Fit

The problem of finding a set of parameters with the best fit can be solved by any available optimization procedure. This includes the on-line (or instance-based) version of the gradient descent method, which corresponds to the well-known delta rule (Rumelhart, Hinton, & Williams, 1986).

Let $f$ denote a parametric value function over the belief space with adjustable weights $\overline{w} = \{w_1, w_2, \cdots, w_k\}$. Then the on-line update for a weight $w_i$ is computed as:

$$w_i \leftarrow w_i - \alpha_i(f(b_j) - y_j)\frac{\partial f}{\partial w_i}|_{b_j},$$

where $\alpha_i$ is a learning constant, and $b_j$ and $y_j$ are the last-seen point and its value. Note that the gradient descent method requires the function to be differentiable with regard to adjustable weights.

To solve the discounted infinite-horizon problem, the stochastic (on-line) version of a least-squares fit can be combined with either parallel (synchronous) or incremental (Gauss-Seidel) point updates. In the first case, the value function from the previous step is fixed and a new value function is computed from scratch using a set of belief point samples and their values computed through one-step expansion. Once the parameters are stabilized (by attenuating learning rates), the newly acquired function is fixed, and the process proceeds with another iteration. In the incremental version, a single value-function model is at the same time updated and used to compute new values at sampled points. Littman et al. (1995) and Parr and Russell (1995) implement this approach using asynchronous reinforcement learning backups in which sample points to be updated next are obtained via stochastic simulation. We stress that all versions are subject to the threat of instability and divergence, as remarked above.

### 4.6.3 Parametric Function Models

To apply the least-squares approach we must first select an appropriate value function model. Examples of simple convex functions are linear or quadratic functions, but more complex models are possible as well.

One interesting and relatively simple approach is based on the least-squares approximation of linear action-value functions (Q-functions) (Littman et al., 1995). Here the value function $\widehat{V}_{i+1}$ is approximated as a piecewise linear and convex combination of $\widehat{Q}_{i+1}$ functions:

$$\widehat{V}_{i+1}(b) = \max_{a \in A} \widehat{Q}_{i+1}(b, a),$$

where $\widehat{Q}_{i+1}(b, a)$ is the least-squares fit of a linear function for a set of sample points $G$. Values at points in $G$ are obtained as

$$\varphi^a_{i+1}(b) = \rho(b, a) + \gamma \sum_{o \in \Theta} P(o|b, a)\widehat{V}_i(\tau(b, o, a)).$$

The method leads to an approximation with $|A|$ linear functions and the coefficients of these functions can be found efficiently by solving a set of linear equations. Recall that other two approximations (the QMDP and the fast informed bound approximations) also work with $|A|$ linear functions. The main differences between the methods are that the QMDP and fast informed bound methods update linear functions directly, and they guarantee upper bounds and unique convergence.

A more sophisticated parametric model of a convex function is the *softmax model* (Parr & Russell, 1995):

$$\widehat{V}(b) = \left[ \sum_{\alpha \in \Gamma} \left[ \sum_{s \in S} \alpha(s) b(s) \right]^k \right]^{\frac{1}{k}},$$

where $\Gamma$ is the set of linear functions $\alpha$ with adaptive parameters to fit and $k$ is a "temperature" parameter that provides a better fit to the underlying piecewise linear convex function for larger values. The function represents a soft approximation of a piecewise linear convex function, with the parameter $k$ smoothing the approximation.

### 4.6.4 CONTROL

We tested the control performance of the least-squares approach on the linear Q-function model (Littman et al., 1995) and the softmax model (Parr & Russell, 1995). For the softmax model we varied the number of linear functions, trying cases with 10 and 15 linear functions respectively. In the first set of experiments we used parallel (synchronous) updates and samples at a fixed set of 100 belief points. We applied stochastic gradient descent techniques to find the best fit in both cases. We tested the control performance for value-function approximations obtained after 10, 20 and 30 updates, starting from the QMDP solution. In the second set of experiments, we applied the incremental stochastic update scheme with Gauss-Seidel-style updates. The results for this method were acquired after every grid point was updated 150 times, with learning rates decreasing linearly in the range between 0.2 and 0.001. Again we started from the QMDP solution. The results for lookahead controllers are summarized in Figure 22, which also shows the control performance of the direct Q-function controller and, for comparison, the results for the QMDP method.

The linear-Q function model performed very well and the results for the lookahead design were better than the results for the QMDP method. The difference was quite apparent for direct approaches. In general, the good performance of the method can be attributed to the choice of a function model that let us match the shape of the optimal value function reasonably well. In contrast, the softmax models (with 10 and 15 linear functions) did not perform as expected. This is probably because in the softmax model all linear functions are updated for every sample point. This leads to situations in which multiple linear functions try to track a belief point during its update. Under these circumstances it is hard to capture the structure of the optimal value function accurately. The other negative feature is that the effects of on-line changes of all linear functions are added in the softmax approximation, and thus could bias incremental update schemes. In the ideal case, we would like to identify one vector $\alpha$ responsible for a specific belief point and update (modify) only that vector. The linear Q-function approach avoids this problem by always updating only a single linear function (corresponding to an action).

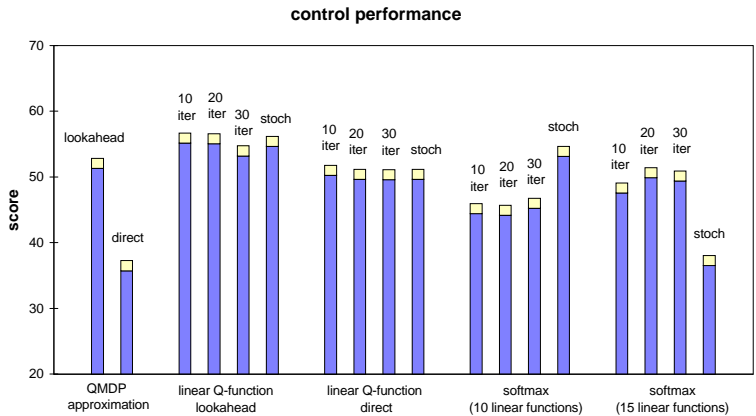

Figure 22: Control performance of least-squares fit methods. Models tested include: linear Q-function model (with both direct and lookahead control) and softmax models with 10 and 15 linear functions (lookahead control only). Value functions obtained after 10, 20 and 30 synchronous updates and value functions obtained through the incremental stochastic update scheme are used to define different controllers. For comparison, we also include results for two QMDP controllers.

## 4.7 Grid-Based Approximations with Linear Function Updates

An alternative grid-based approximation method can be constructed by applying Sondik's approach for computing derivatives (linear functions) to points of the grid (Lovejoy, 1991a, 1993). Let $\widehat{V}_i$ be a piecewise linear convex function described by a set of linear functions $\Gamma_i$. Then a new linear function for a belief point $b$ and an action $a$ can be computed efficiently as (Smallwood & Sondik, 1973; Littman, 1996)

$$\alpha_{i+1}^{b,a}(s) = \rho(s,a) + \gamma \sum_{o \in \Theta} \sum_{s' \in S} P(s',o|s,a) \alpha_i^{\iota(b,a,o)}(s'), \qquad (10)$$

where $\iota(b,a,o)$ indexes a linear function $\alpha_i$ in a set of linear functions $\Gamma_i$ (defining $\widehat{V}_i$) that maximizes the expression

$$\sum_{s' \in S} \left[ \sum_{s \in S} P(s',o|s,a)b(s) \right] \alpha_i(s')$$

for a fixed combination of $b, a, o$. The optimizing function for $b$ is then acquired by choosing the vector with the best overall value from all action vectors. That is, assuming $\Gamma_{i+1}^b$ is a set of all candidate linear functions, the resulting functions satisfies

$$\alpha_{i+1}^{b,*} = \arg \max_{\alpha_{i+1}^b \in \Gamma_{i+1}^b} \sum_{s \in S} \alpha_{i+1}^b(s)b(s).$$

A collection of linear functions obtained for a set of belief points can be combined into a piecewise linear and convex value function. This is the idea behind a number of exact algorithms (see Section 2.4.2). However, in the exact case, a set of points that cover all linear functions defining the new value function must be located first, which is a hard task in itself. In contrast, the approximation method uses an incomplete set of belief points that are fixed or at least easy to locate, for example via random or heuristic selection. We use $H_{GL}$ to denote the value-function mapping for the grid approach.

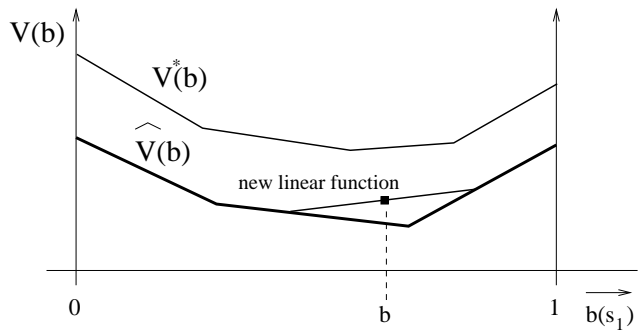

Figure 23: An incremental version of the grid-based linear function method. The piecewise linear lower bound is improved by a new linear function computed for a belief point $b$ using Sondik's method.

The advantage of the grid-based method is that it leads to more efficient updates. The time complexity of the update is polynomial and equals $O(|G||A||S|^2|\Theta|)$. It yields a set of $|G|$ linear functions, compared to $|A||\Gamma_i|^{|\Theta|}$ possible functions for the exact update.

Since the set of grid-points is incomplete, the resulting approximation lower-bounds the value function one would obtain by performing the exact update (Lovejoy, 1991a).

**Theorem 13** (*Lower-bound property of the grid-based linear function update*). *Let $\widehat{V}_i$ be a piecewise linear value function and $G$ a set of grid points used to compute linear function updates. Then $H_{GL}\widehat{V}_i \leq H\widehat{V}_i$.*

### 4.7.1 INCREMENTAL LINEAR-FUNCTION APPROACH

The drawback of the grid-based linear function method is that $H_{GL}$ is not a contraction for the discounted infinite-horizon case, and therefore the value iteration method based on the mapping may not converge (Lovejoy, 1991a). To remedy this problem, we propose an incremental version of the grid-based linear function method. The idea of this refinement is to prevent instability by gradually improving the piecewise linear and convex lower bound of the value function.

Assume that $\widehat{V}_i \leq V^*$ is a convex piecewise linear lower bound of the optimal value function defined by a linear function set $\Gamma_i$, and let $\alpha_b$ be a linear function for a point $b$ that is computed from $\widehat{V}_i$ using Sondik's method. Then one can construct a new improved value function $\widehat{V}_{i+1} \geq \widehat{V}_i$ by simply adding the new linear function $\alpha^b$ into $\Gamma_i$. That is: $\Gamma_{i+1} = \Gamma_i \cup \alpha_b$. The idea of the incremental update, illustrated in Figure 23, is similar to incremental methods used by Cheng (1988) and Lovejoy (1993). The method can be extended to handle a set of grid points $G$ in a straightforward way. Note also that after adding one or more new linear functions to $\Gamma_i$, some of the previous linear functions may become redundant and can be removed from the value function. Techniques for redundancy checking are the same as are applied in the exact approaches (Monahan, 1982; Eagle, 1984).

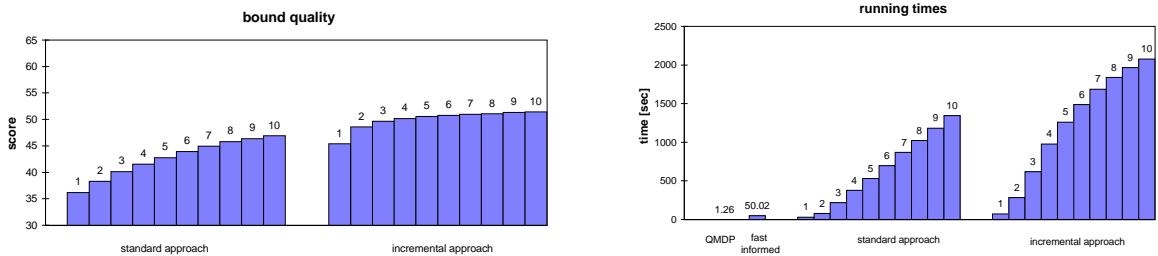

Figure 24: Bound quality and running times of the standard and incremental version of the grid-based linear-function method for the fixed 40-point grid. Cumulative running times (including all previous update cycles) are shown for both methods. Running times of the QMDP and the fast informed bound methods are included for comparison.

The incremental refinement is stable and converges for a fixed set of grid points. The price paid for this feature is that the linear function set $\Gamma_i$ can grow in size over the iteration steps. Although the growth is at most linear in the number of iterations, compared to the potentially exponential growth of exact methods, the linear function set describing the piecewise linear approximation can become huge. Thus, in practice we usually stop incremental updates well before the method converges. The question that remains open is the complexity (hardness) of the problem of finding the fixed-point solution for a fixed set of grid points $G$.

Figure 24 illustrates some of the trade-offs involved in applying incremental updates compared to the standard fixed-grid approach on the Maze20 problem. We use the same grid of 40 points for both techniques and the same initial value function. Results for 1-10 update cycles are shown. We see that the incremental method has longer running times than the standard method, since the number of linear functions can grow after every update. On the other hand, the bound quality of the incremental method improves more rapidly and it can never become worse after more update steps.

### 4.7.2 MINIMUM EXPECTED REWARD

The incremental method improves the lower bound of the value function. The value function, say $\widehat{V}_i$, can be used to create a controller (with either the lookahead or direct-action choice). In the general case, we cannot say anything about the performance quality of such controllers with regard to $\widehat{V}_i$. However, under certain conditions the performance of both controllers is guaranteed never to fall below $\widehat{V}_i$. The following theorem (proved in the Appendix) establishes these conditions.

**Theorem 14** *Let $\widehat{V}_i$ be a value function obtained via the incremental linear function method, starting from $\widehat{V}_0$, which corresponds to some fixed strategy $C_0$. Let $C_{LA,i}$ and $C_{DR,i}$ be two* *controllers based on $\widehat{V}_i$: the lookahead controller and the direct action controller, and $V^{C_{LA,i}}$, $V^{C_{DR,i}}$ be their respective value functions. Then $\widehat{V}_i \le V^{C_{LA,i}}$ and $\widehat{V}_i \le V^{C_{DR,i}}$ hold.*

We note that the same property holds for the incremental version of exact value iteration. That is, both the lookahead and the direct controllers perform no worse than $V_i$ obtained after $i$ incremental updates from some $V_0$ corresponding to a FSM controller $C_0$.

### 4.7.3 SELECTING GRID POINTS

The incremental version of the grid-based linear-function approximation is flexible and works for an arbitrary grid.[26] Moreover, the grid need not be fixed and can be changed on line. Thus, the problem of finding grids reduces to the problem of selecting belief points to be updated next. One can apply various strategies to do this. For example, one can use a fixed set of grid points and update them repeatedly, or one can select belief points on line using various heuristics.

The incremental linear function method guarantees that the value function is always improved (all linear functions from previous steps are kept unless found to be redundant). The quality of a new linear function (to be added next) depends strongly on the quality of linear functions obtained in previous steps. Therefore, our objective is to select and order points with better chances of larger improvement. To do this we have designed two heuristic strategies for selecting and ordering belief points.

The first strategy attempts to optimize updates at extreme points of the belief simplex by ordering them heuristically. The idea of the heuristic is based on the fact that states with higher expected rewards (e.g. some designated goal states) backpropagate their effects (rewards) locally. Therefore, it is desirable that states in the neighborhood of the highest reward state be updated first, and the distant ones later. We apply this idea to order extreme points of the belief simplex, relying on the current estimate of the value function to identify the highest expected reward states and on a POMDP model to determine the neighbor states.

The second strategy is based on the idea of stochastic simulation. The strategy generates a sequence of belief points more likely to be reached from some (fixed) initial belief point. The points of the sequence are then used in reverse order to generate updates. The intent of this heuristic is to "maximize" the improvement of the value function at the initial fixed point. To run this heuristic, we need to find an initial belief point or a set of initial belief points. To address this problem, we use the first heuristic that allows us to order the extreme points of the belief simplex. These points are then used as initial beliefs for the simulation part. Thus, we have a two-tier strategy: the top-level strategy orders extremes of the belief simplex, and the lower-level strategy applies stochastic simulation to generate a sequence of belief states more likely reachable from a specific extreme point.

We tested the order heuristics and the two-tier heuristics on our Maze20 problem, and compared them also to two simple point selection strategies: the fixed-grid strategy, in which a set of 40 grid points was updated repeatedly, and the random-grid strategy, in which points were always chosen uniformly at random. Figure 25 shows the bound quality of the methods for 10 update cycles (each cycle consists of 40 grid point updates) on the Maze20 problem. We see that the differences in the quality of value-function approximations for different strategies (even the very simple ones) are relatively small. We note that we observed similar results also for other problems, not just Maze20.

---

26. There is no restriction on the grid points that must be included in the grid, such as was required for example in the linear point-interpolation scheme, which had to use all extreme points of the belief simplex.

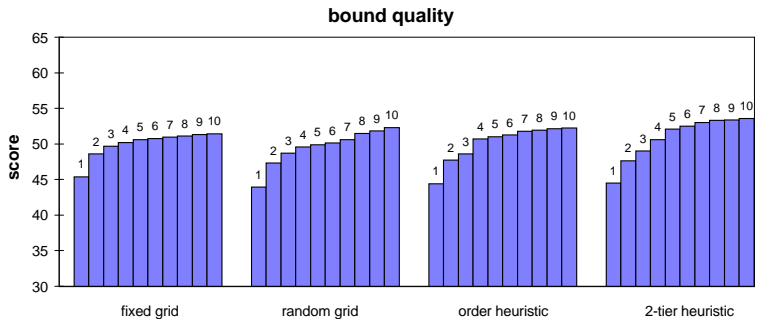

Figure 25: Improvements in the bound quality for the incremental linear-function method and four different grid-selection heuristics. Each cycle includes 40 grid-point updates.

The relatively small improvement of our heuristics can be explained by the fact that every new linear function influences a larger portion of the belief space and thus the method should be less sensitive to a choice of a specific point.[27] However, another plausible explanation is that our heuristics were not very good and more accurate heuristics or combinations of heuristics could be constructed. Efficient strategies for locating grid points used in some of the exact methods, e.g. the Witness algorithm (Kaelbling et al., 1999) or Cheng's methods (Cheng, 1988) can potentially be applied to this problem. This remains an open area of research.

### 4.7.4 CONTROL

The grid-based linear-function approach leads to a piecewise linear and convex approximation. Every linear function comes with a natural action choice that lets us choose the action greedily. Thus we can run both the lookahead and the direct controllers. Figure 26 compares the performance of four different controllers for the fixed grid of 40 points, combining standard and incremental updates with lookahead and direct greedy control after 1, 5 and 10 update cycles. The results (see also Figure 24) illustrate the trade-offs between the computational time of obtaining the solution and its quality. We see that the incremental approach and the lookahead controller design tend to improve the control performance. The prices paid are worse running and reaction times, respectively.

---

27. The small sensitivity of the incremental method to the selection of grid points would suggest that one could, in many instances, replace exact updates with simpler point selection strategies. This could increase the speed of exact value-iteration methods (at least in their initial stages), which suffer from inefficiencies associated with locating a complete set of grid points to be updated in every step. However, this issue needs to be investigated.

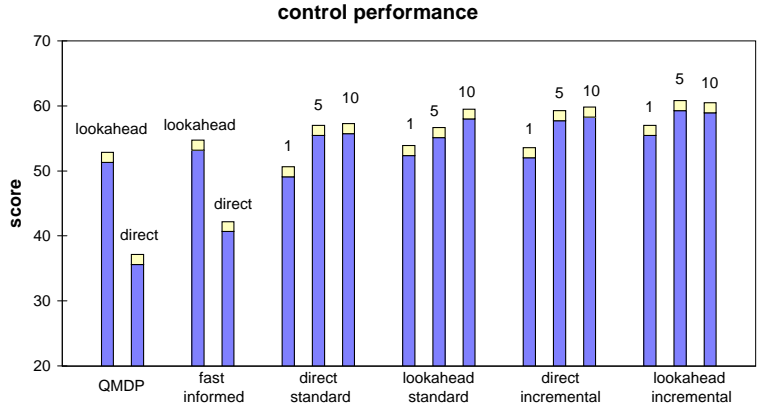

Figure 26: Control performance of four different controllers based on grid-based linear function updates after 1, 5 and 10 update cycles for the same 40-point grid. Controllers represent combinations of two update strategies (standard and incremental) and two action-extraction techniques (direct and lookahead). Running times for the two update strategies were presented in Figure 24. For comparison we include also performances of the QMDP and the fast informed bound methods (with both direct and lookahead designs).

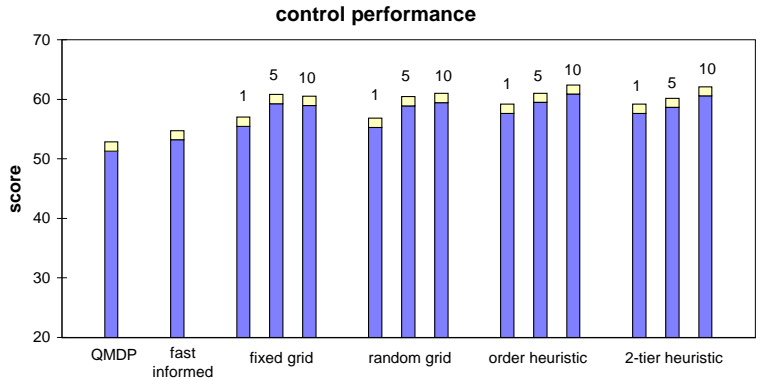

Figure 27: Control performances of lookahead controllers based on the incremental linear-function approach and different point-selection heuristics after 1, 5 and 10 improvement cycles. For comparison, scores for the QMDP and the fast informed bound approximations are shown as well.

Figure 27 illustrates the effect of point selection heuristics on control. We compare the results for lookahead control only, using approximations obtained after 1, 5 and 10 improvement cycles (each cycle consists of 40 grid point updates). The test results show that, as for the bound quality, there are no big differences among various heuristics, suggesting a small sensitivity of control to the selection of grid points.

## 4.8 Summary of Value-Function Approximations

Heuristic value-function approximations methods allow us to replace hard-to-compute exact methods and trade off solution quality for speed. There are numerous methods we can employ, each with different properties and different trade-offs of quality versus speed. Tables 1 and 2 summarize main theoretical properties of the approximation methods covered in this paper. The majority of these methods are of polynomial complexity or at least have efficient (polynomial) Bellman updates. This makes them good candidates for more complex POMDP problems that are out of reach of exact methods.

All of the methods are heuristic approximations in that they do not give solutions of a guaranteed precision. Despite this fact we proved that solutions of some of the methods are no worse than others in terms of value function quality (see Figure 15). This was one of the main contributions of the paper. However, there are currently minimal theoretical results relating these methods in terms of control performance; the exception are some results for FSM-controllers and FSM-based approximations. The key observation here is that for the quality of control (lookahead control) it is more important to approximate the shape (derivatives) of the value function correctly. This is also illustrated empirically on grid-based interpolation-extrapolation methods in Section 4.5.9 that are based on non-convex value functions. The main challenges here are to find ways of analyzing and comparing control performance of different approximations also theoretically and to identify classes of POMDPs for which certain methods dominate the others.

Finally, we note that the list of methods is not complete and other value-function approximation methods or the refinements of existing methods are possible. For example, White and Scherer (1994) investigate methods based on truncated histories that lead to upper and lower bound estimates of the value function for complete information states (complete histories). Also, additional restrictions on some of the methods can change the properties of a more generic method. For example, it is possible that under additional assumptions we will be able to ensure convergence of the least-squares fit approximation.

## 5. Conclusions

POMDPs offers an elegant mathematical framework for representing decision processes in stochastic partially observable domains. Despite their modeling advantages, however, POMDP problems are hard to solve exactly. Thus, the complexity of problem solving-procedures becomes the key aspect in the sucessful application of the model to real-world problems, even at the expense of the optimality. As recent complexity results for the approximability of POMDP problems are not encouraging (Lusena et al., 1998; Madani et al., 1999), we focus on heuristic approximations, in particular approximations of value functions.

| Method | Bound | Isotonicity | Contraction |
|---|---|---|---|
| MDP approximation | upper | √ | √ |
| QMDP approximation | upper | √ | √ |
| Fast informed bound | upper | √ | √ |
| UMDP approximation | lower | √ | √ |
| Fixed-strategy method | lower | √ | √ |
| Grid-based interpolation-extrapolation | - | - | - |
|    Nearest neighbor | - | √ | √ |
|    Kernel regression | - | √ | √ |
|    Linear point interpolation | upper | √ | √ |
| Curve-fitting (least-squares fit) | - | - | - |
|    linear Q-function | - | - | - |
| Grid-based linear function method | lower | - | - |
|    Incremental version (start from a lower bound) | lower | √ | - * |

Table 1: Properties of different value-function approximation methods: bound property, isotonicity and contraction property of the underlying mappings for $0 \leq \gamma < 1$. (*) Although incremental version of the grid-based linear-function method is not a contraction it always converges.

| Method | Finite-horizon | Discounted infinite-horizon |
|---|---|---|
| MDP approximation | P | P |
| QMDP approximation | P | P |
| Fast informed bound | P | P |
| UMDP approximation | NP-hard | undecidable |
| Fixed-strategy method | P | P |
| Grid-based interpolation-extrapolation | varies | NA |
|    Nearest neighbor | P | P |
|    Kernel regression | P | P |
|    Linear point interpolation | P | varies |
|      Fixed interpolation | P | P |
|      Best interpolation | P | ? |
| Curve-fitting (least-squares fit) | varies | NA |
|    linear Q-function | P | NA |
| Grid-based linear function method | P | NA |
|    Incremental version | NA | ? |

Table 2: Complexity of value-function approximation methods for finite-horizon problem and discounted infinite-horizon problem. The objective for the discounted infinite-horizon case is to find the corresponding fixed-point solution. The complexity results take into account, in addition to components of POMDPs, also all other approximation specific parameters, e.g., the size of the grid $G$ in grid-based methods. ? indicates open instances and NA methods that are not applicable to one of the problems (e.g. because of possible divergence).

## 5.1 Contributions

The paper surveys new and known value-function approximation methods for solving POMDPs. We focus primarily on the theoretical analysis and comparison of the methods, with findings and results supported experimentally on a problem of moderate size from the agent navigation domain. We analyze the methods from different perspectives: their computational complexity, capability to bound the optimal value function, convergence properties of iterative implementations, and the quality of derived controllers. The analysis includes new theoretical results, deriving the properties of individual approximations, and their relations to exact methods. In general, the relations between and trade-offs among different methods are not well understood. We provide some new insights on these issues by analyzing their corresponding updates. For example, we showed that the differences among the exact, the MDP, the QMDP, the fast-informed bound, and the UMDP methods boil down to simple mathematical manipulations and their subsequent effect on the value-function approximation. This allowed us to determine relations among different methods in terms of quality of their respective value functions which is one of the main results of the paper.

We also presented a number of new methods and heuristic refinements of some existing techniques. The primary contributions in this area include the fast-informed bound, grid-based point interpolation methods (including adaptive grid approaches based on stochastic sampling), and the incremental linear-function method. We also showed that in some instances the solutions can be obtained more efficiently by converting the original approximation into an equivalent finite-state MDP. For example, grid-based approximations with convex rules can be often solved via conversion into a grid-based MDP (in which grid points correspond to new states), leading to the polynomial-complexity algorithm for both the finite and the discounted infinite-horizon cases (Section 4.5.3). This result can dramatically improve the run-time performance of the grid-based approaches. A similar conversion to the equivalent finite-state MDP, allowing a polynomial-time solution for the discounted infinite-horizon problem, was shown for the fast informed bound method (Section 4.2).

## 5.2 Challenges and Future Directions

Work on POMDPs and their approximations is far from complete. Some complexity results remain open, in particular, the complexity of the grid-based approach seeking the best interpolation, or the complexity of finding the fixed-point solution for the incremental version of the grid-based linear-function method. Another interesting issue that needs more investigation is the convergence of value iteration with least-squares approximation. Although the method can be unstable in the general case, it is possible that under certain restrictions it will converge.

In the paper we use a single POMDP problem (Maze20) only to support theoretical findings or to illustrate some intuitions. Therefore, the results not supported theoretically (related mostly to control) cannot be generalized and used to rank different methods, since their performance may vary on other problems. In general, the area of POMDPs and POMDP approximations suffers from a shortage of larger-scale experimental work with multiple problems of different complexities and a broad range of methods. Experimental work is especially needed to study and compare different methods with regard to control quality. The main reason for this is that there are only few theoretical results relating the control performance. These studies should help focus theoretical exploration by discovering interesting cases and possibly identifying classes of problems for which certain approximations are more or less suitable. Our preliminary experimental results show that there are significant differences in control performance among different methods and that not all of them may be suitable to approximate the control policies. For example, the grid-based nearest-neighbor approach with piecewise-constant approximation is typically inferior to and outperformed by other simpler (and more efficient) value-function methods.

The present work focused on heuristic approximation methods. We investigated general (flat) POMDPs and did not take advantage of any additional structural refinements. However, real-world problems usually offer more structure that can be exploited to devise new algorithms and perhaps lead to further speed-ups. It is also possible that some of the restricted versions of POMDPs (with additional structural assumptions) can be solved or approximated efficiently, even though the general complexity results for POMDPs or their $\epsilon$-approximations are not very encouraging (Papadimitriou & Tsitsiklis, 1987; Littman, 1996; Mundhenk et al., 1997; Lusena et al., 1998; Madani et al., 1999). A challenge here is to identify models that allow efficient solutions and are at the same time interesting enough from the point of application.

Finally, a number of interesting issues arise when we move to problems with large state, action, and observation spaces. Here, the complexity of not only value-function updates but also belief state updates becomes an issue. In general, partial observability of hidden process states does not allow us to factor and decompose belief states (and their updates), even when transitions have a great deal of structure and can be represented very compactly. Promising directions to deal with these issues include various Monte-Carlo approaches (Isard & Blake, 1996; Kanazawa, Koller, & Russell, 1995; Doucet, 1998; Kearns et al., 1999)), methods for approximating belief states via decomposition (Boyen & Koller, 1998, 1999), or a combination of the two approaches (McAllester & Singh, 1999).

## Acknowledgements

Anthony Cassandra, Thomas Dean, Leslie Kaelbling, William Long, Peter Szolovits and anonymous reviewers provided valuable feedback and comments on this work. This research was supported by grant RO1 LM 04493 and grant 1T15LM07092 from the National Library of Medicine, by DOD Advanced Research Project Agency (ARPA) under contract number N66001-95-M-1089 and DARPA/Rome Labs Planning Initiative grant F30602-95-1-0020.

## Appendix A. Theorems and proofs

### A.1 Convergence to the Bound

**Theorem 6** *Let $H_1$ and $H_2$ be two value-function mappings defined on $\mathcal{V}_1$ and $\mathcal{V}_2$ s.t.*

1. *$H_1$, $H_2$ are contractions with fixed points $V_1^*$, $V_2^*$;*

2. *$V_1^* \in \mathcal{V}_2$ and $H_2 V_1^* \geq H_1 V_1^* = V_1^*$;*

3. *$H_2$ is an isotone mapping.*

*Then $V_2^* \geq V_1^*$ holds.*

**Proof** By applying $H_2$ to condition 2 and expanding the result with condition 2 again we get: $H_2^2 V_1^* \geq H_2 V_1^* \geq H_1 V_1^* = V_1^*$. Repeating this we get in the limit $V_2^* \geq \cdots \geq H_2^n V_1^* \geq \cdots H_2^2 V_1^* \geq H_2 V_1^* \geq H_1 V_1^* = V_1^*$, which proves the result. □

## A.2 Accuracy of a Lookahead Controller Based on Bounds

**Theorem 7** Let $\widehat{V}_U$ and $\widehat{V}_L$ be upper and lower bounds of the optimal value function for the discounted infinite-horizon problem. Let $\epsilon = \sup_b |\widehat{V}_U(b) - \widehat{V}_L(b)| = \|\widehat{V}_U - \widehat{V}_L\|$ be the maximum bound difference. Then the expected reward for a lookahead controller $\widehat{V}^{LA}$, constructed for either $\widehat{V}_U$ or $\widehat{V}_L$, satisfies $\|\widehat{V}^{LA} - V^*\| \leq \frac{\epsilon(2-\gamma)}{(1-\gamma)}$.

**Proof** Let $\widehat{V}$ denotes either an upper or lower bound approximation of $V^*$ and $H^{LA}$ be the value function mapping corresponding to the lookahead policy for $\widehat{V}$. Note, that since the lookahead policy always optimizes its actions with regard to $\widehat{V}$, $H\widehat{V} = H^{LA}\widehat{V}$ must hold. The error of $\widehat{V}^{LA}$ can be bounded using the triangle inequality

$$\|\widehat{V}^{LA} - V^*\| \leq \|\widehat{V}^{LA} - \widehat{V}\| + \|\widehat{V} - V^*\|.$$

The first component satisfies:

$$\begin{aligned}
\|\widehat{V}^{LA} - \widehat{V}\| &= \|H^{LA}\widehat{V}^{LA} - \widehat{V}\| \\
&\leq \|H^{LA}\widehat{V}^{LA} - H\widehat{V}\| + \|H\widehat{V} - \widehat{V}\| \\
&= \|H^{LA}\widehat{V}^{LA} - H^{LA}\widehat{V}\| + \|H\widehat{V} - \widehat{V}\| \\
&\leq \gamma\|\widehat{V}^{LA} - \widehat{V}\| + \epsilon
\end{aligned}$$

The inequality: $\|H\widehat{V} - \widehat{V}\| \leq \epsilon$ follows from the isotonicity of $H$ and the fact that $\widehat{V}$ is either an upper or a lower bound. Rearranging the inequalities, we obtain: $\|\widehat{V}^{LA} - \widehat{V}\| = \frac{\epsilon}{(1-\gamma)}$. The bound on the second term $\|\widehat{V} - V^*\| \leq \epsilon$ is trivial.

Therefore, $\|\widehat{V}^{LA} - V^*\| \leq \epsilon[\frac{1}{(1-\gamma)} + 1] = \epsilon\frac{(2-\gamma)}{(1-\gamma)}$. □

## A.3 MDP, QMDP and the Fast Informed Bounds

**Theorem 8** A solution for the fast informed bound approximation can be found by solving an MDP with $|S||A||\Theta|$ states, $|A|$ actions and the same discount factor $\gamma$.

**Proof** Let $\alpha_i^a$ be a linear function for action $a$ defining $\widehat{V}_i$. Let $\alpha_i(s,a)$ denote parameters of the function. The parameters of $\widehat{V}_{i+1}$ satisfy:

$$\alpha_{i+1}(s,a) = \rho(s,a) + \gamma \sum_{o \in \Theta} \max_{a' \in A} \sum_{s' \in S} P(s',o|s,a)\alpha_i(s',a').$$

Let

$$\alpha_{i+1}(s,a,o) = \max_{a' \in A} \sum_{s' \in S} P(s',o|s,a)\alpha_i(s',a').$$

87

Now, we can rewrite $\alpha_{i+1}(s, a, o)$ for every $s, a, o$ as:

$$
\begin{aligned}
\alpha_{i+1}(s, a, o) &= \max_{a' \in A} \left\{ \sum_{s' \in S} P(s', o | s, a) \left[ \rho(s', a') + \gamma \sum_{o' \in \Theta} \alpha_i(s', a', o') \right] \right\} \\
&= \max_{a' \in A} \left\{ \left[ \sum_{s' \in S} P(s', o | s, a) \rho(s', a') \right] + \gamma \left[ \sum_{o' \in \Theta} \sum_{s' \in S} P(s', o | s, a) \alpha_i(s', a', o') \right] \right\}
\end{aligned}
$$

These equations define an MDP with state space $S \times A \times \Theta$, action space $A$ and discount factor $\gamma$. Thus, a solution for the fast informed bound update can be found by solving an equivalent finite-state MDP.  □

**Theorem 9** *Let $\widehat{V}_i$ corresponds to a piecewise linear convex value function defined by $\Gamma_i$ linear functions. Then $H\widehat{V}_i \leq H_{FIB}\widehat{V}_i \leq H_{QMDP}\widehat{V}_i \leq H_{MDP}\widehat{V}_i$.*

**Proof**

$$
\begin{aligned}
& \max_{a \in A} \left\{ \sum_{s \in S} \rho(s, a) b(s) + \gamma \sum_{o \in \Theta} \max_{\alpha_i \in \Gamma_i} \sum_{s' \in S} \sum_{s \in S} P(s', o | s, a) b(s) \alpha_i(s') \right\} \\
= \ & (HV_i)(b) \\
\leq \ & \max_{a \in A} \sum_{s \in S} b(s) \left[ \rho(s, a) + \gamma \sum_{o \in \Theta} \max_{\alpha_i \in \Gamma_i} \sum_{s' \in S} P(s', o | s, a) \alpha_i(s') \right] \\
= \ & (H_{FIB} V_i)(b) \\
\leq \ & \max_{a \in A} \sum_{s \in S} b(s) \left[ \rho(s, a) + \gamma \sum_{s' \in S} P(s' | s, a) \max_{\alpha_i \in \Gamma_i} \alpha_i(s') \right] \\
= \ & (H_{QMDP} \widehat{V}_i)(b) \\
\leq \ & \sum_{s \in S} b(s) \max_{a \in A} \left[ \rho(s, a) + \gamma \sum_{s' \in S} P(s' | s, a) \max_{\alpha_i \in \Gamma_i} \alpha_i(s') \right] \\
= \ & (H_{MDP} \widehat{V}_i)(b) \quad \square
\end{aligned}
$$

## A.4 Fixed-Strategy Approximations

**Theorem 10** *Let $C_{FSM}$ be an FSM controller. Let $C_{DR}$ and $C_{LA}$ be the direct and the one-step-lookahead controllers constructed based on $C_{FSM}$. Then $V^{C_{FSM}}(b) \leq V^{C_{DR}}(b)$ and $V^{C_{FSM}}(b) \leq V^{C_{LA}}(b)$ hold for all belief states $b \in \mathcal{I}$.*

**Proof** The value function for the FSM controller $C_{FSM}$ satisfies:

$$
V^{C_{FSM}}(b) = \max_{x \in M} V(x, b) = V(\psi(b), b)
$$

where

$$
V(x, b) = \rho(b, \eta(x)) + \gamma \sum_{o \in \Theta} P(o | b, \eta(x)) V(\phi(x, o), \tau(b, \eta(x), o)).
$$

The direct controller $C_{DR}$ selects the action greedily in every step, that is, it always chooses according to $\psi(b) = \arg\max_{x \in M} V(x, b)$. The lookahead controller $C_{LA}$ selects the action based on $V(x, b)$ one step away:

$$\eta^{LA}(b) = \arg\max_{a \in A} \left[ \rho(b, a) + \gamma \sum_{o \in \Theta} P(o|b, a) \max_{x' \in M} V(x', \tau(b, a, o)) \right].$$

By expanding the value function for $C_{FSM}$ for one step we get:

$$V^{C_{FSM}}(b) = \max_{x \in M} V(x, b)$$

$$= \max_{x \in M} \left[ \rho(b, \eta(x)) + \gamma \sum_{o \in \Theta} P(o|b, \eta(x)) V(\phi(x, o), \tau(b, \eta(x), o)) \right] \quad (1)$$

$$= \rho(b, \eta(\psi(b))) + \gamma \sum_{o \in \Theta} P(o|b, \eta(\psi(b))) V(\phi(x, o), \tau(b, \eta(\psi(b)), o))$$

$$\leq \rho(b, \eta(\psi(b))) + \gamma \sum_{o \in \Theta} P(o|b, \eta(\psi(b))) \max_{x' \in M} V(x', \tau(b, \eta(\psi(b)), o)) \quad (2)$$

$$\leq \max_{a \in A} \left[ \rho(b, a) + \gamma \sum_{o \in \Theta} P(o|b, a) \max_{x' \in M} V(x', \tau(b, a, o)) \right]$$

$$= \rho(b, \eta^{LA}(b)) + \gamma \sum_{o \in \Theta} P(o|b, \eta^{LA}(b)) \max_{x' \in M} V(x', \tau(b, \eta^{LA}(b), o)) \quad (3)$$

Iteratively expanding $\max_{x' \in M} V(x, .)$ in 2 and 3 with expression 1 and substituting improved (higher value) expressions 2 and 3 back we obtain value functions for both the direct and the lookahead controllers. (Expansions of 2 lead to the value for the direct controller and expansions of 3 to the value for the lookahead controller.) Thus $V^{C_{FSM}} \leq V^{C_{DR}}$ and $V^{C_{FSM}} \leq V^{C_{LA}}$ must hold. Note, however, that action choices $\psi(b)$ and $\psi^{LA}(b)$ in expressions 2 and 3 can be different leading to different next step belief states and subsequently to different expansion sequences. Therefore, the above result does not imply that $V^{DR}(b) \leq V^{LA}(b)$ for all $b \in \mathcal{I}$. $\quad \square$

## A.5 Grid-Based Linear-Function Method

**Theorem 14** *Let $\widehat{V}_i$ be a value function obtained via the incremental linear function method, starting from $\widehat{V}_0$, which corresponds to some fixed strategy $C_0$. Let $C_{LA,i}$ and $C_{DR,i}$ be two controllers based on $\widehat{V}_i$: the lookahead controller and the direct action controller, and $V^{C_{LA,i}}$, $V^{C_{DR,i}}$ be their respective value functions. Then $\widehat{V}_i \leq V^{C_{LA,i}}$ and $\widehat{V}_i \leq V^{C_{DR,i}}$ hold.*

**Proof** By initializing the method with a value function for some FSM controller $C_0$, the incremental updates can be interpreted as additions of new states to the FSM controller (a new linear function corresponds to a new state of the FSM). Let $C_i$ be a controller after step $i$. Then $V^{C_{FSM,i}} = \widehat{V}_i$ holds and the inequalities follow from Theorem 10. $\quad \square$

# References

Astrom, K. J. (1965). Optimal control of Markov decision processes with incomplete state estimation. *Journal of Mathematical Analysis and Applications*, *10*, 174–205.

Baird, L. C. (1995). Residual algorithms: Reinforcement learning with function approximation. In *Proceedings of the Twelfth International Conference on Machine Learning*, pp. 30–37.

Barto, A. G., Bradtke, S. J., & Singh, S. P. (1995). Learning to act using real-time dynamic programming. *Artificial Intelligence*, *72*, 81–138.

Bellman, R. E. (1957). *Dynamic programming*. Princeton University Press, Princeton, NJ.

Bertsekas, D. P. (1994). A counter-example to temporal differences learning. *Neural Computation*, *7*, 270–279.

Bertsekas, D. P. (1995). *Dynamic programming and optimal control*. Athena Scientific.

Bonet, B., & Geffner, H. (1998). Learning sorting and classification with POMDPs. In *Proceedings of the Fifteenth International Conference on Machine Learning*.

Boutilier, C., Dean, T., & Hanks, S. (1999). Decision-theoretic planning: Structural assumptions and computational leverage. *Artificial Intelligence*, *11*, 1–94.

Boutilier, C., & Poole, D. (1996). Exploiting structure in policy construction. In *Proceedings of the Thirteenth National Conference on Artificial Intelligence*, pp. 1168–1175.

Boyan, J. A., & Moore, A. A. (1995). Generalization in reinforcement learning: safely approximating the value function. In *Advances in Neural Information Processing Systems 7*. MIT Press.

Boyen, X., & Koller, D. (1998). Tractable inference for complex stochastic processes. In *Proceedings of the Fourteenth Conference on Uncertainty in Artificial Intelligence*, pp. 33–42.

Boyen, X., & Koller, D. (1999). Exploiting the architecture of dynamic systems. In *Proceedings of the Sixteenth National Conference on Artificial Intelligence*, pp. 313–320.

Brafman, R. I. (1997). A heuristic variable grid solution method for POMDPs. In *Proceedings of the Fourteenth National Conference on Artificial Intelligence*, pp. 727–233.

Burago, D., Rougemont, M. D., & Slissenko, A. (1996). On the complexity of partially observed Markov decision processes. *Theoretical Computer Science*, *157*, 161–183.

Cassandra, A. R. (1998). *Exact and approximate algorithms for partially observable Markov decision processes*. Ph.D. thesis, Brown University.

Cassandra, A. R., Littman, M. L., & Zhang, N. L. (1997). Incremental pruning: a simple, fast, exact algorithm for partially observable Markov decision processes. In *Proceedings of the Thirteenth Conference on Uncertainty in Artificial Intelligence*, pp. 54–61.

Castañon, D. (1997). Approximate dynamic programming for sensor management. In *Proceedings of Conference on Decision and Control*.

Cheng, H.-T. (1988). *Algorithms for partially observable Markov decision processes*. Ph.D. thesis, University of British Columbia.

Condon, A. (1992). The complexity of stochastic games. *Information and Computation*, *96*, 203–224.

Dean, T., & Kanazawa, K. (1989). A model for reasoning about persistence and causation. *Computational Intelligence*, *5*, 142–150.

Dearden, R., & Boutilier, C. (1997). Abstraction and approximate decision theoretic planning. *Artificial Intelligence*, *89*, 219–283.

Doucet, A. (1998). On sequential simulation-based methods for Bayesian filtering. Tech. rep. CUED/F-INFENG/TR 310, Department of Engineering, Cambridge University.

Drake, A. (1962). *Observation of a Markov process through a noisy channel*. Ph.D. thesis, Massachusetts Institute of Technology.

Draper, D., Hanks, S., & Weld, D. (1994). Probabilistic planning with information gathering and contingent execution. In *Proceedings of the Second International Conference on AI Planning Systems*, pp. 31–36.

Eagle, J. N. (1984). The optimal search for a moving target when search path is constrained. *Operations Research*, *32*, 1107–1115.

Eaves, B. (1984). *A course in triangulations for soving differential equations with deformations*. Springer-Verlag, Berlin.

Gordon, G. J. (1995). Stable function approximation in dynamic programming. In *Proceedings of the Twelfth International Conference on Machine Learning*.

Hansen, E. (1998a). An improved policy iteration algorithm for partially observable MDPs. In *Advances in Neural Information Processing Systems 10*. MIT Press.

Hansen, E. (1998b). Solving POMDPs by searching in policy space. In *Proceedings of the Fourteenth Conference on Uncertainty in Artificial Intelligence*, pp. 211–219.

Hauskrecht, M. (1997). *Planning and control in stochastic domains with imperfect information*. Ph.D. thesis, Massachusetts Institute of Technology.

Hauskrecht, M., & Fraser, H. (1998). Planning medical therapy using partially observable Markov decision processes. In *Proceedings of the Ninth International Workshop on Principles of Diagnosis (DX-98)*, pp. 182–189.

Hauskrecht, M., & Fraser, H. (2000). Planning treatment of ischemic heart disease with partially observable Markov decision processes. *Artificial Intelligence in Medicine*, *18*, 221–244.

Heyman, D., & Sobel, M. (1984). *Stochastic methods in operations research: stochastic optimization*. McGraw-Hill.

Howard, R. A. (1960). *Dynamic Programming and Markov Processes*. MIT Press, Cambridge.

Howard, R. A., & Matheson, J. (1984). Influence diagrams. *Principles and Applications of Decision Analysis, 2*.

Isard, M., & Blake, A. (1996). Contour tracking by stochastic propagation of conditional density. In *Proccedings of Europian Conference on Computer Vision*, pp. 343–356.

Kaelbling, L. P., Littman, M. L., & Cassandra, A. R. (1999). Planning and acting in partially observable stochastic domains. *Artificial Intelligence, 101*, 99–134.

Kanazawa, K., Koller, D., & Russell, S. J. (1995). Stochastic simulation algorithms for dynamic probabilistic networks. In *Proceedings of the Eleventh Conference on Uncertainty in Artificial Intelligence*, pp. 346–351.

Kearns, M., Mansour, Y., & Ng, A. Y. (1999). A sparse sampling algorithm for near optimal planning in large Markov decision processes. In *Proceedings of the Sixteenth International Joint Conference on Artificial Intelligence*, pp. 1324–1331.

Kjaerulff, U. (1992). A computational scheme for reasoning in dynamic probabilistic networks. In *Proceedings of the Eighth Conference on Uncertainty in Artificial Intelligence*, pp. 121–129.

Korf, R. (1985). Depth-first iterative deepening: an optimal admissible tree search. *Artificial Intelligence, 27*, 97–109.

Kushmerick, N., Hanks, S., & Weld, D. (1995). An algorithm for probabilistic planning. *Artificial Intelligence, 76*, 239–286.

Lauritzen, S. L. (1996). *Graphical models*. Clarendon Press.

Littman, M. L. (1994). Memoryless policies: Theoretical limitations and practical results. In Cliff, D., Husbands, P., Meyer, J., & Wilson, S. (Eds.), *From Animals to Animats 3: Proceedings of the Third International Conference on Simulation of Adaptive Behavior*. MIT Press, Cambridge.

Littman, M. L. (1996). *Algorithms for sequential decision making*. Ph.D. thesis, Brown University.

Littman, M. L., Cassandra, A. R., & Kaelbling, L. P. (1995). Learning policies for partially observable environments: scaling up. In *Proceedings of the Twelfth International Conference on Machine Learning*, pp. 362–370.

Lovejoy, W. S. (1991a). Computationally feasible bounds for partially observed Markov decision processes. *Operations Research, 39*, 192–175.

Lovejoy, W. S. (1991b). A survey of algorithmic methods for partially observed Markov decision processes. *Annals of Operations Research, 28*, 47–66.

Lovejoy, W. S. (1993). Suboptimal policies with bounds for parameter adaptive decision processes. *Operations Research, 41*, 583–599.

Lusena, C., Goldsmith, J., & Mundhenk, M. (1998). Nonapproximability results for Markov decision processes. Tech. rep., University of Kentucky.

Madani, O., Hanks, S., & Condon, A. (1999). On the undecidability of probabilistic planning and infinite-horizon partially observable Markov decision processes. In *Proceedings of the Sixteenth National Conference on Artificial Intelligence*.

McAllester, D., & Singh, S. P. (1999). Approximate planning for factored POMDPs using belief state simplification. In *Proceedings of the Fifteenth Conference on Uncertainty in Artificial Intelligence*, pp. 409–416.

McCallum, R. (1995). Instance-based utile distinctions for reinforcement learning with hidden state. In *Proceedings of the Twelfth International Conference on Machine Learning*.

Monahan, G. E. (1982). A survey of partially observable Markov decision processes: theory, models, and algorithms. *Management Science, 28*, 1–16.

Mundhenk, M., Goldsmith, J., Lusena, C., & Allender, E. (1997). Encyclopaedia of complexity results for finite-horizon Markov decision process problems. Tech. rep., CS Dept TR 273-97, University of Kentucky.

Papadimitriou, C. H., & Tsitsiklis, J. N. (1987). The complexity of Markov decision processes. *Mathematics of Operations Research, 12*, 441–450.

Parr, R., & Russell, S. (1995). Approximating optimal policies for partially observable stochastic domains. In *Proceedings of the Fourteenth International Joint Conference on Artificial Intelligence*, pp. 1088–1094.

Pearl, J. (1988). *Probabilistic reasoning in intelligent systems*. Morgan Kaufman.

Platzman, L. K. (1977). *Finite memory estimation and control of finite probabilistic systems*. Ph.D. thesis, Massachusetts Institute of Technology.

Platzman, L. K. (1980). A feasible computational approach to infinite-horizon partially-observed Markov decision problems. Tech. rep., Georgia Institute of Technology.

Puterman, M. L. (1994). *Markov decision processes: discrete stochastic dynamic programming*. John Wiley, New York.

Raiffa, H. (1970). *Decision analysis. Introductory lectures on choices under uncertainty*. Addison-Wesley.

Rumelhart, D., Hinton, G. E., & Williams, R. J. (1986). Learning internal representations by error propagation. In *Parallel Distributed Processing*, pp. 318–362.

Satia, J., & Lave, R. (1973). Markovian decision processes with probabilistic observation of states. *Management Science, 20,* 1–13.

Singh, S. P., Jaakkola, T., & Jordan, M. I. (1994). Learning without state-estimation in partially observable Markovian decision processes. In *Proceedings of the Eleventh International Conference on Machine Learning*, pp. 284–292.

Smallwood, R. D., & Sondik, E. J. (1973). The optimal control of partially observable processes over a finite horizon. *Operations Research, 21,* 1071–1088.

Sondik, E. J. (1971). *The optimal control of partially observable Markov decision processes.* Ph.D. thesis, Stanford University.

Sondik, E. J. (1978). The optimal control of partially observable processes over the infinite horizon: Discounted costs. *Operations Research, 26,* 282–304.

Tatman, J., & Schachter, R. D. (1990). Dynamic programming and influence diagrams. *IEEE Transactions on Systems, Man and Cybernetics, 20,* 365–379.

Tsitsiklis, J. N., & Roy, B. V. (1996). Feature-based methods for large-scale dynamic programming. *Machine Learning, 22,* 59–94.

Washington, R. (1996). Incremental Markov model planning. In *Proceedings of the Eight IEEE International Conference on Tools with Artificial Intelligence*, pp. 41–47.

White, C. C., & Scherer, W. T. (1994). Finite memory suboptimal design for partially observed Markov decision processes. *Operations Research, 42,* 439–455.

Williams, R. J., & Baird, L. C. (1994). Tight performance bounds on greedy policies based on imperfect value functions. In *Proceedings of the Tenth Yale Workshop on Adaptive and Learning Systems* Yale University.

Yost, K. A. (1998). *Solution of large-scale allocation problems with partially observable outcomes.* Ph.D. thesis, Naval Postgraduate School, Monterey, CA.

Zhang, N. L., & Lee, S. S. (1998). Planning with partially observable Markov decision processes: Advances in exact solution method. In *Proceedings of the Fourteenth Conference on Uncertainty in Artificial Intelligence*, pp. 523–530.

Zhang, N. L., & Liu, W. (1997a). A model approximation scheme for planning in partially observable stochastic domains. *Journal of Artificial Intelligence Research, 7,* 199–230.

Zhang, N. L., & Liu, W. (1997b). Region-based approximations for planning in stochastic domains. In *Proceedings of the Thirteenth Conference on Uncertainty in Artificial Intelligence*, pp. 472–480.